\documentclass[10pt,journal,compsoc]{IEEEtran}
\ifCLASSOPTIONcompsoc
  \usepackage[nocompress]{cite}
\else
  \usepackage{cite}
\fi
\usepackage[dvips]{graphicx}
\ifCLASSINFOpdf
\else
\fi
\usepackage{bbm}
\usepackage{amsmath, amssymb,bm}
\usepackage{amsfonts}
\usepackage{subfigure}
\usepackage{algorithm}
\usepackage{algpseudocode}

\usepackage{booktabs}

\usepackage{xcolor}

\begin{document}
%
\title{Aggregated Multi-output Gaussian Processes
with Knowledge Transfer Across Domains}
\author{Yusuke~Tanaka,
        Toshiyuki~Tanaka,~\IEEEmembership{Member,~IEEE,}
	Tomoharu~Iwata,
	Takeshi~Kurashima,\\
	Maya~Okawa,
	Yasunori~Akagi,
	Hiroyuki~Toda
	\IEEEcompsocitemizethanks{\IEEEcompsocthanksitem Yusuke Tanaka is with the NTT
	Communication Science Laboratories, NTT Corporation, Kyoto, 619-0237, Japan.\protect\\
	E-mail: yusuke.tanaka.rh@hco.ntt.co.jp}
	\thanks{Manuscript received April 19, 2005; revised August 26, 2015.}
	}

%
%

\markboth{Journal of \LaTeX\ Class Files,~Vol.~14, No.~8, August~2015}%
{Shell \MakeLowercase{\textit{et al.}}: Bare Demo of IEEEtran.cls for Computer Society Journals}
%



\IEEEtitleabstractindextext{%
\begin{abstract}
Aggregate data often appear in various fields
such as socio-economics and public security.
The aggregate data are associated
not with points but with {\em supports}
(e.g., spatial regions in a city).
Since the supports may have various granularities
depending on {\em attributes} (e.g., poverty rate and crime rate),
modeling such data is not straightforward.
This article offers a multi-output Gaussian process (MoGP) model
that infers functions for attributes
using multiple aggregate datasets of respective granularities.
In the proposed model,
the function for each attribute is assumed to be
a dependent GP modeled as a linear mixing of independent latent GPs.
We design an observation model with an aggregation process for each attribute;
the process is an integral of the GP over the corresponding support.
We also introduce a prior distribution of the mixing weights,
which allows a knowledge transfer across {\em domains} (e.g., cities)
by sharing the prior.
This is advantageous in such a situation where
the spatially aggregated dataset in a city is too coarse to interpolate;
the proposed model can still make accurate predictions of attributes
by utilizing aggregate datasets in other cities.
The inference of the proposed model is based on variational Bayes,
which enables one to learn the model parameters
using the aggregate datasets from multiple domains.
The experiments demonstrate that the proposed model outperforms
in the task of refining coarse-grained aggregate data
on real-world datasets: Time series of air pollutants in Beijing
and various kinds of spatial datasets from New York City and Chicago.
\end{abstract}

\begin{IEEEkeywords}
Gaussian processes, aggregate data, variational Bayes
\end{IEEEkeywords}}

\maketitle

\IEEEdisplaynontitleabstractindextext

%
\IEEEpeerreviewmaketitle

\IEEEraisesectionheading{\section{Introduction}\label{sec:introduction}}

\begin{figure}[!t]
 \centering
 \includegraphics[width=88mm, bb=0 0 2568 2330]{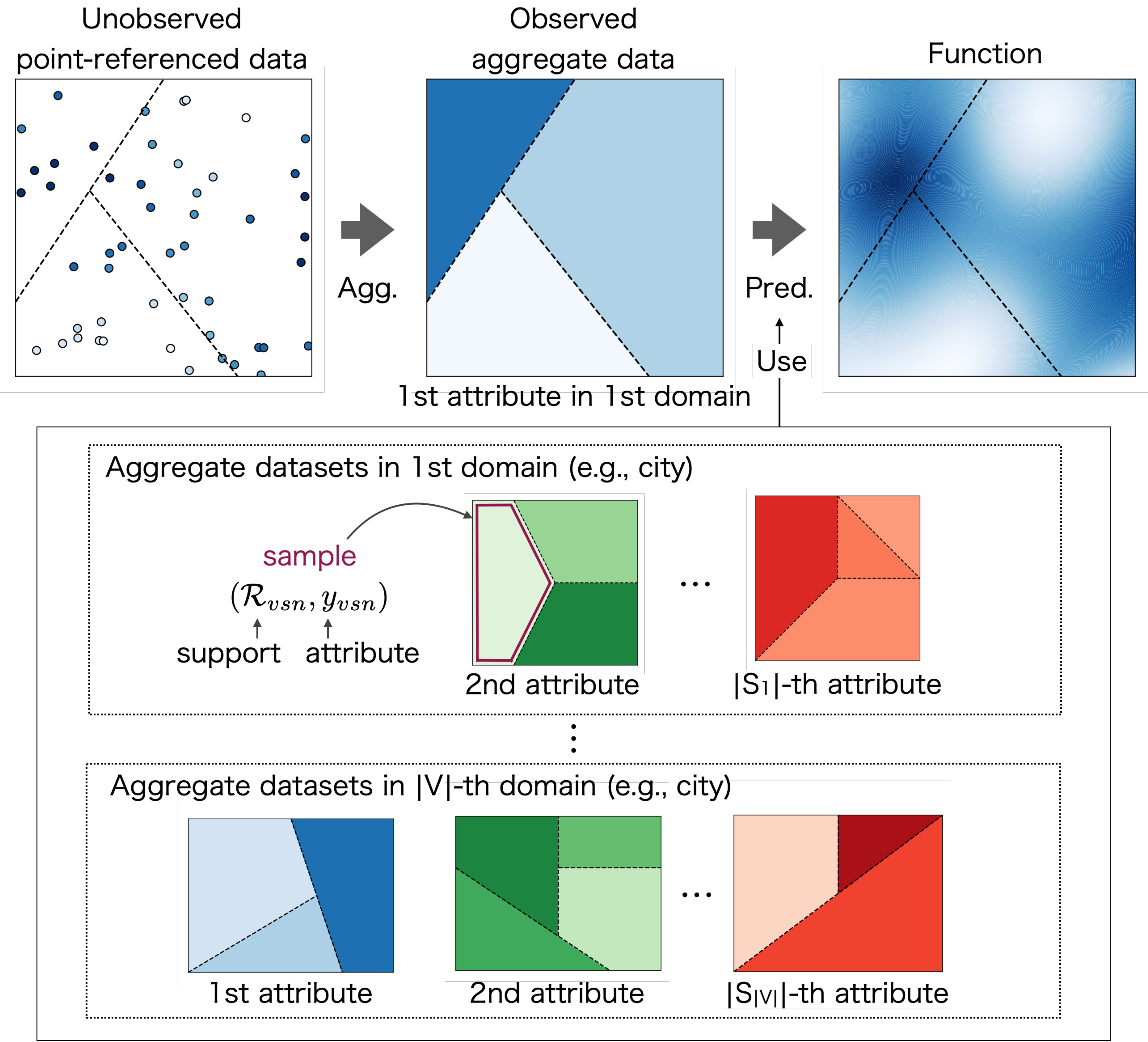}
 \caption{The problem setting when using aggregate datasets
 defined on two-dimensional domains.
 Darker hues represent higher attribute values.
 Assume that we obtain aggregate datasets in multiple domains,
 where each attribute value is given by aggregating
 point-referenced data over the corresponding support.
 Note that we do not use point-referenced data
 in either training or test phases.
 The goal is to infer a function for the attribute
 using aggregate datasets in multiple domains.
 }
 \label{fig:problem}
\end{figure}

%
%
%
%
\IEEEPARstart{A}{ggregate} data often appear in various fields
such as socio-economics~\cite{rupasinghaa:social,Smith:poverty},
public security~\cite{bogomolov:once,wang:crime},
ecology~\cite{keil:downscaling},
agricultural economics~\cite{howitt:spatial,xavier:disaggregating},
epidemiology~\cite{sturrock:fine},
meteorology~\cite{wilby:guidelines,zorita:analog},
public health~\cite{jerrett:spatial},
urban planning~\cite{tanaka:time-delayed},
and remote sensing~\cite{yousefi:multi}.
The aggregate data contain a pair of {\em support} and {\em attribute}.
The support is a predefined unit for aggregation,
such as a time bin and a spatial region.
The attribute value is obtained by aggregating point-referenced data
over the corresponding support (see Figure~\ref{fig:problem}).
For example, time series of air pollutant concentration
gathered by low-cost sensors is associated with
coarse-grained time bins (e.g., six hours)
because the point-referenced data are averaged
over the bin to alleviate possible noise effects.
Another example is the city's poverty rate collected via household surveys;
the point-referenced data are aggregated over spatial regions (e.g., districts)
to preserve privacy.

In this article, we suppose the situation
where various aggregate datasets in multiple domains
are available and consider the problem of inferring functions
for respective attributes.
Here, the {\em domain} indicates an input space;
for example, the domain is one-dimensional when we are interested
in time series, and it is two-dimensional when we consider
spatial data obtained from a city. 
The problem setting for two-dimensional domains
is illustrated in Figure~\ref{fig:problem}.
The estimated function can be used to make predictions of attributes at any point,
which is significant in various applications,
such as improving city environments~\cite{Smith:poverty,wang:crime}.
For instance, analyzing the spatial distribution of poverty in a city
allows one to optimize the allocation of resources for remedial action.

This problem is challenging
because the aggregated attributes are obtained at
supports with various granularities,
namely different shapes and sizes
(see Figure~\ref{fig:problem}).
Learning is difficult, especially when the data are sparse,
that is, they are associated with coarse-grained supports.
In that case, one promising approach is joint modeling of all attributes;
however, it is still not obvious how to establish dependences between
aggregated attributes across multiple domains.

Gaussian processes (GPs) are nonparametric distributions over functions,
which are widely used as priors to infer unknown functions from
data~\cite{carl:gaussian}.
Multi-output Gaussian processes (MoGPs) are promising
to tackle the data sparsity issue,
allowing one to learn functions by considering dependences between
attributes~\cite{alvarez:kernels}.
However, almost all the GP-based models assume that
the samples are obtained at points;
they are not straightforwardly applicable to aggregate data observed at
supports~\cite{bonilla:multi-task,teh:semiparametric,boyle:dependent,nguyen:collaborative,yu:learning,micchelli:kernels,higdon:space,luttinen:variational}.

This article presents a probabilistic model,
called {\em Aggregated Multi-output Gaussian Processes (A-MoGPs)},
to infer functions for attributes
using aggregate datasets in multiple domains.
In A-MoGP, the functions for attributes are assumed to be
a dependent GP modeled as a linear mixing of independent latent GPs.
The covariance functions of the latent GPs
are shared among attributes and domains.
By introducing a prior distribution of the mixing weights shared among domains,
one can obtain appropriate estimates of
the covariance functions and the weights
via a knowledge transfer mechanism,
even if some aggregate datasets have coarse granularities.
The critical component of A-MoGP is
to have an observation model with aggregation processes,
in which attribute values are assumed to be calculated
by integrating the mixed GP over the corresponding support.
The covariances between supports are then given by
the double integral of the covariance function
over the corresponding pair of supports.
This component is helpful because one can accurately evaluate
the covariance between supports considering support shape and size.

The inference of A-MoGP is based on variational Bayes (VB).
The model parameters can be learned
by maximizing the evidence lower bound (ELBO),
in which GPs are analytically integrated out.
We adopt the reparameterization trick~\cite{kingma:auto},
which allows us to use gradient-based optimization methods
for learning the variational parameters.
By deriving the predictive distribution,
we can obtain the functions for respective attributes
considering the covariance of data points
and the dependences between aggregated attributes simultaneously.

The main contributions of this article are as follows~\footnote{
A preliminary version of this work appeared
in the Proceedings of NeurIPS'19~\cite{tanaka:spatially}.
The main differences of this article from~\cite{tanaka:spatially}
are as follows.
We introduce the prior of the weight parameters
and develop the learning algorithm based on variational Bayes,
allowing one to utilize aggregate datasets gathered
from multiple domains via the knowledge transfer.
We also conduct extensive experiments
on real-world aggregate datasets;
we especially add the evaluations
using aggregated time series of pollutant concentrations.}:
\begin{itemize} 
 \item We propose A-MoGP,
       a novel multi-output GP model that incorporates
       the aggregation process for each attribute in multiple domains.
 \item We develop a VB algorithm 
       that can learn model parameters by maximizing the ELBO,
       in which latent GPs are analytically integrated out.
       This is the first derivation of the predictive distribution
       given aggregate datasets in multiple domains.
 \item The experiments on real-world datasets
       defined in temporal or spatial domains
       demonstrate the effectiveness of A-MoGP
       in the task of refining coarse-grained aggregate data.
\end{itemize}

This article is organized as follows:
In Section~\ref{sec:related}, we describe related works.
Section~\ref{sec:aggregate} describes aggregate datasets in multiple domains.
In Section~\ref{sec:model}, we propose A-MoGP
for inferring functions from aggregate datasets in multiple domains.
In Section~\ref{sec:inference},
we present the VB algorithm for learning model parameters
and derive the predictive distribution.
Section~\ref{sec:experiments} demonstrates the effectiveness of A-MoGP
using multiple real-world aggregate datasets.
Finally, we describe concluding remarks
and a discussion of future work in Section~\ref{sec:conclusion}.

\section{Related work}
\label{sec:related}
The problem of refining spatially aggregated data
has long been addressed in the geostatistics community
under the name of {\em statistical downscaling}, {\em spatial disaggregation},
and {\em areal interpolation};
the problem of predicting point-referenced data
from aggregate data is also called
the {\em change of support} problem~\cite{gotway:combining}.
One difficulty in these problems is that
the covariance of aggregate data
is not equal to that of point-referenced data,
which is called the ecological fallacy~\cite{robinson:ecological,King:solution}
in the field of statistics.
In order to estimate the covariance from aggregate data precisely,
it has been indicated that data aggregation processes
should be incorporated into the models,
as in prior works (e.g.,~\cite{gotway:combining,park:spatial})
as well as in our proposal.
The following paragraphs describe existing methods
for addressing the problem we focus on,
which can be roughly categorized into two approaches:
Regression approach and multivariate approach.

A regression approach has been adopted frequently
for refining coarse-grained aggregate data.
This approach distinguishes multiple datasets
into one target dataset and the others (auxiliary datasets)
and then models the target attribute
as a linear or non-linear mapping of the auxiliary
attributes~\cite{murakami:new,park:spatial,keil:downscaling,taylor:continuous,wilson:pointless}.
These models have the aggregation process
only for the target attribute,
encouraging consistency between the fine- and coarse-grained target attribute.
The aggregation process is incorporated via block kriging~\cite{burgess:optimal}
or transformations of Gaussian process (GP)
priors~\cite{smith:transformations,smith:gaussian}.
In recent years, regression-based models for aggregate settings
have been studied in the field of machine
learning~\cite{Law:variational,tanaka:refining,yivan:learning,siu:deconditional}.
It has been pointed out that
this task is a kind of multiple instance
learning~\cite{Law:variational,yivan:learning}.
The regression-based models assume that
all the auxiliary datasets have
sufficiently fine granularities
(e.g., 5 minutes intervals in the timeline
and 1~km $\times$ 1~km grid cells in the geographical space);
thus they do not consider aggregation constraints
for the auxiliary datasets.
However, this assumption is not always fulfilled;
for example, spatially aggregated datasets from cities
are often associated with various geographical partitions
such as districts and police precincts;
hence, one might not be able to
access fine-grained auxiliary datasets.
In such cases, the regression-based approach
cannot fully use all the aggregate datasets
containing the coarse-grained ones.

An alternative for modeling multiple datasets is
a multivariate approach.
Unlike the regression approach,
this approach does not distinguish multiple datasets;
it aims to design a joint distribution of all attributes.
Generally, the multivariate approach is expected to
alleviate data sparsity issues.
Multi-output Gaussian processes (MoGPs)~\cite{alvarez:kernels}
and co-kriging~\cite{myers:matrix} are typical choices
for modeling multivariate data
that can consider covariances of data points
and dependences between attributes simultaneously.
Along the research line of MoGPs,
there have been several sophisticated methods,
including process convolution~\cite{boyle:dependent,higdon:space}
and latent factor modeling~\cite{bonilla:multi-task,teh:semiparametric,nguyen:collaborative,yu:learning,micchelli:kernels,luttinen:variational},
for establishing dependences between attributes.
The linear model of coregionalization (LMC) is
the general and most widely-used framework
for constructing a multivariate function,
in which outputs (i.e., attributes) are represented as
a linear combination of independent latent
functions~\cite{pierre:geostatistics,andre:mining}.
The MoGP models based on latent factor modeling are instances of LMC,
where the latent functions are defined by GPs.
Unfortunately, all the existing MoGP models
cannot straightforwardly apply to
the aggregate data this study focuses on
because they assume point-referenced data.
In other words, they do not have an important mechanism,
that is, the aggregation constraints,
for handling attributes aggregated over supports.

The proposed model is an extension of LMC.
To handle the aggregate data,
we introduce an observation model
with the aggregation process for all attributes;
this is represented by the integral of the MoGP
over each corresponding support, as in~\cite{smith:gaussian}.
We also present the variational Bayes algorithm
for learning model parameters
and derive the predictive distribution,
given aggregate datasets in multiple domains.
MoGP models for aggregate data
have been independently proposed at the same
time~\cite{hamelijnck:multi,yousefi:multi,tanaka:spatially}.
The main differences of ours from~\cite{hamelijnck:multi,yousefi:multi}
are as follows:
(a) derivation of the predictive distribution
given aggregate datasets from multiple domains;
(b) knowledge transfer across domains
by incorporating the prior distribution of mixing weights;
(c) extensive experiments on real-world aggregate datasets
gathered from temporal or spatial domains.

\section{Aggregate Data}
\label{sec:aggregate}
In this section,
we introduce mathematical notations of
aggregate datasets in multiple domains.
Let $\mathcal{V}$ be a set of domain indices.
Let $\mathcal{X}_v \subset \mathbb{R}^D (v\in\mathcal{V})$ be
a domain of dimension $D$.
In the case of $D=1$,
a typical example of the domain is an observation time period;
a domain example in the case of $D=2$ is a total region of a city.
When we consider multiple domains, i.e., $|\mathcal{V}|\ge2$, 
where $|\bullet|$ is the cardinality of the set,
we regard that different domains do not intersect. 
Let $\bm{x}\in\mathcal{X}_v (v\in\mathcal{V})$ denote an input variable.
Let $\mathcal{S}_v (v\in\mathcal{V})$ be a set of attribute indices
for each domain.
A partition $\mathcal{P}_{vs} (v\in\mathcal{V},s\in\mathcal{S}_v)$
of $\mathcal{X}_v$ is a collection of disjoint subsets,
called {\it supports}, of $\mathcal{X}_v$.
Let $\mathcal{R}_{vsn}\in\mathcal{P}_{vs}$ be an $n$-th support
in $\mathcal{P}_{vs}$.
The support corresponds to the time bin ($D=1$),
the spatial region ($D=2$), and so on.
A data sample is specified by a pair $(\mathcal{R}_{vsn},y_{vsn})$,
where $y_{vsn}\in\mathbb{R}$ is an attribute value
that is observed by aggregating the point-referenced data
over the corresponding support $\mathcal{R}_{vsn}$ (see Figure~\ref{fig:problem}).
Here, we assume that the aggregation process (e.g., averaging)
for each attribute is known.
Suppose that we have the collection of aggregate datasets
$\mathcal{D}=\{\mathcal{D}_v \mid v\in\mathcal{V}\}$,
where $\mathcal{D}_v=\{(\mathcal{R}_{vsn},y_{vsn})\mid 
s\in\mathcal{S}_v;n=1,\ldots,|\mathcal{P}_{vs}|\}$.
The notations used in this article are listed in Table~\ref{tb:notation}.

\begin{table}[!t]
 \centering
 \caption{Notation of aggregate datasets.}
 \begin{tabular}{ l l } \toprule
  Symbol&Description  \\ \midrule
  $\mathcal{V}$ &set of domain indices \\[2pt]
  $v$ &domain index, $v\in\mathcal{V}$ \\[2pt]
  $D$ &dimension of domain, $D\in\mathbb{N}$ \\[2pt]
  $\mathcal{X}_v$ &$v$-th domain, $\mathcal{X}_v\subset\mathbb{R}^D$ \\[2pt]
  $\mathcal{S}_v$ &set of attribute indices for $v$-th domain \\[2pt]
  $s$ &attribute index, $s\in\mathcal{S}_v$ \\[2pt]
  $\mathcal{P}_{vs}$ &partition of $\mathcal{X}_v$ for $s$-th attribute \\[2pt]
  $n$ &support index, $n\in\{1,\ldots,|\mathcal{P}_{vs}|\}$ \\[2pt]
  $\mathcal{R}_{vsn}$ &support, $\mathcal{R}_{vsn}\in\mathcal{P}_{vs}$
      \\[2pt]
  $y_{vsn}$ &attribute value associated with the support
      $\mathcal{R}_{vsn}$ \\[2pt]
  \bottomrule
 \end{tabular}
 \label{tb:notation}
\end{table}

\section{Model}
\label{sec:model}
\begin{figure}[!t]
 \centering
 \includegraphics[width=80mm, bb=0 0 1742 2262]{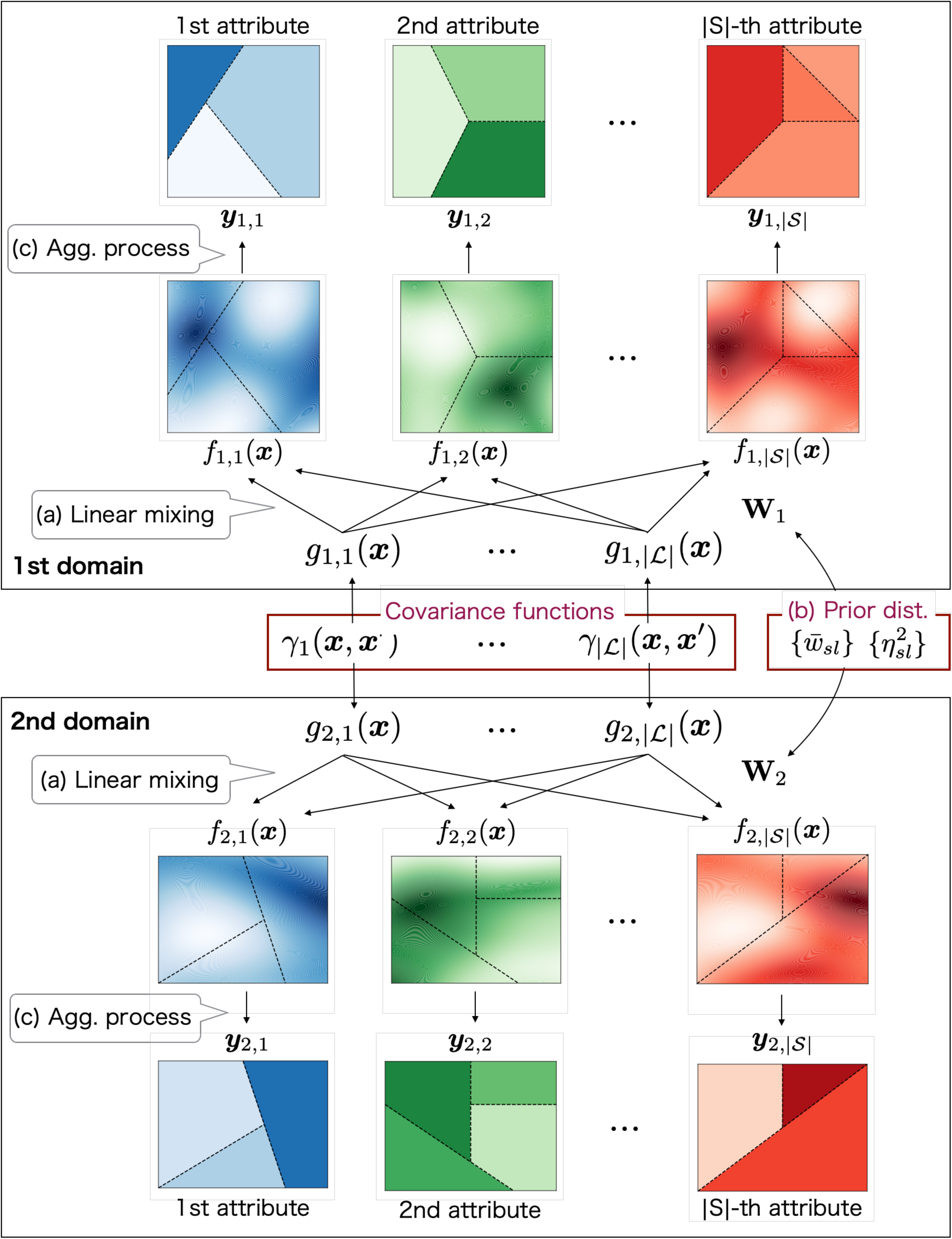}
 \caption{Schematic diagram of A-MoGP:
 Generative process of multiple aggregated attributes in two spatial domains.
 Covariance functions and prior distributions of weight parameters
 are shared among domains.}
 \label{fig:model}
\end{figure}

We propose A-MoGP (Aggregated Multi-output Gaussian Process),
a probabilistic model for inferring functions
from aggregate datasets with various granularities in multiple domains.
The proposed model when we have two spatial domains
is illustrated schematically in Figure~\ref{fig:model}.
For simplicity, we assume that the types of attributes
are the same for different domains;
namely, we let $\mathcal{S}=\mathcal{S}_v$ denote 
the set of attribute indices for all domains. 
Notice that one can straightforwardly apply the proposed model
to the case where the types of attributes are
partially different across domains.
In the following,
we first construct a multi-output Gaussian process (MoGP) prior
by linearly mixing independent latent GPs
to establish dependences between attributes
(see (a) Linear mixing in Figure~\ref{fig:model}).
We then introduce a prior distribution of the mixing weights
to transfer knowledge across domains
(see (b) Prior dist. in Figure~\ref{fig:model}).
Lastly, we present an observation model with aggregation processes
for respective attributes
(see (c) Agg. process in Figure~\ref{fig:model}),
allowing a model learning from aggregate datasets.

{\bf MoGP prior.}
In the proposed model,
the functions for attributes
on the continuous space
are assumed to be an MoGP.
We construct the MoGP by linearly mixing
some independent latent GPs,
which is one of the most widely used approaches
for establishing dependences between
outputs (i.e., attributes)~\cite{teh:semiparametric,alvarez:kernels}.
Let $\mathcal{L}$ be an index set of latent GPs.
Consider $|\mathcal{V}||\mathcal{L}|$ independent zero-mean GPs,
\begin{flalign}
 g_{vl}(\bm{x})
 &\sim \mathcal{GP}\left(
 0, \gamma_l(\bm{x}, \bm{x}') 
 \right),
 \quad v\in\mathcal{V};\, l\in \mathcal{L},
 \label{eq:latent_GP}
\end{flalign}
where $\gamma_l(\bm{x}, \bm{x}'):
\mathbb{R}^D \times \mathbb{R}^D\to\mathbb{R}$
is a covariance function
for the $l$-th latent GP,
which is assumed integrable.
It should be noted that
the covariance functions
$\{\gamma_l(\bm{x},\bm{x}')\mid l\in\mathcal{L}\}$
are shared among all attributes and all domains,
which enables us to effectively learn
covariances of data points
by utilizing dependences between attributes in multiple domains.
Defining $f_{vs}(\bm{x})$ as the GP
for the $s$-th attribute in the $v$-th domain,
the $|\mathcal{S}|$-dimensional dependent GP
$\bm{f}_v(\bm{x})=(f_{v1}(\bm{x}),\ldots,f_{v|\mathcal{S}|}(\bm{x}))^\top$
in the $v$-th domain
is assumed to be modeled as a linear combination of
the $|\mathcal{L}|$ independent latent GPs.
The MoGP for the $v$-th domain is then given by
\begin{flalign}
 \bm{f}_v(\bm{x})
 &= {\bf W}_v \bm{g}_v(\bm{x}),
 \label{eq:product_wg}
\end{flalign}
where $\bm{g}_v(\bm{x})=
(g_{v1}(\bm{x}),\ldots,g_{v|\mathcal{L}|}(\bm{x}))^\top$,
and where ${\bf W}_v$ is an $|\mathcal{S}| \times |\mathcal{L}|$
weight matrix whose $(s,l)$-entry $w_{vsl} \in \mathbb{R}$
is the weight of the $l$-th latent GP in the $s$-th attribute.
Since a linear combination of GPs is again a GP,
$\bm{f}_v(\bm{x})$ can be written by
\begin{flalign}
 \bm{f}_v(\bm{x}) \mid {\bf W}_v
 &\sim \mathcal{GP}
 \left(
 \bm{0}, {\bf K}_v(\bm{x}, \bm{x}')
 \right),
 \label{eq:MOGP}
\end{flalign}
where $\bm{0}$ is a column vector of 0's,
and where 
${\bf K}_v(\bm{x}, \bm{x}'): \mathcal{X}_v \times \mathcal{X}_v
\to\mathbb{R}^{|\mathcal{S}|\times|\mathcal{S}|}$
is the matrix-valued covariance function represented by
\begin{flalign}
 {\bf K}_v(\bm{x}, \bm{x}')
 &= {\bf W}_v {\bf \Gamma}(\bm{x}, \bm{x}') {\bf W}_v^\top.
 \label{eq:kernel}
\end{flalign}
Here, 
${\bf \Gamma}(\bm{x}, \bm{x}') =
{\rm diag} \left( \gamma_1(\bm{x}, \bm{x}'),\ldots,
\gamma_{|\mathcal{L}|}(\bm{x}, \bm{x}') \right)$.
The $(s,s')$-entry of ${\bf K}_v(\bm{x}, \bm{x}')$ is given by
\begin{flalign}
 k_{vss'}(\bm{x}, \bm{x}')
 &= \sum_{l=1}^L w_{vsl} w_{vs'l} \gamma_l(\bm{x}, \bm{x}').
 \label{eq:kernel_elem}
\end{flalign}
From~\eqref{eq:kernel_elem},
one can see that
the covariance functions
${\bf \Gamma}(\bm{x}, \bm{x}')$ for latent GPs
are shared by all attributes and all domains.
In this article,
we focus on the case $|\mathcal{L}|\leq|\mathcal{V}||\mathcal{S}|$,
with the aim of reducing the model complexity
as this is helpful in alleviating over-fitting,
as in~\cite{teh:semiparametric}.

{\bf Prior of the weights.}
Each weight $w_{vsl}$ is assumed to be generated
from a Gaussian prior distribution,
\begin{flalign}
 w_{vsl}
 &\sim \mathcal{N}\left(
 w_{vsl} \mid
 \bar{w}_{sl},\eta_{sl}^2
 \right),
 \label{eq:prior_w}
\end{flalign}
where $\bar{w}_{sl}$ and $\eta_{sl}^2$ are
a mean and a variance, respectively.
$\eta_{sl}^2$ controls the degrees of knowledge transfer across domains.
When $\eta_{sl}^2$ is close to zero,
the weights $\{w_{vsl}\}_{v\in\mathcal{V}}$ for all domains
are likely to take the same value.
This allows the model to learn parameters
from the full use of all datasets
by appropriately transferring knowledge between domains.
This modeling is especially beneficial
when some datasets in a domain are too coarse
to learn the dependences between attributes.

{\bf Observation model for aggregate data.}
To handle aggregate data,
we design an observation model
with aggregation processes
that are integrals of GPs over corresponding supports.
Let $\bm{y}_{vs} =(y_{vs1},\ldots,y_{vs|\mathcal{P}_{vs}|})$ be
a $|\mathcal{P}_{vs}|$-dimensional vector
consisting of the values
for the $s$-th attribute in the $v$-th domain.
Let $\bm{y}_v=(\bm{y}_{v1},\ldots,\bm{y}_{v|\mathcal{S}|})^\top$ denote 
an $N_v$-dimensional vector consisting of the values
for all attributes in the $v$-th domain,
where $N_v=\sum_{s\in\mathcal{S}} |\mathcal{P}_{vs}|$ is
the total number of samples in the $v$-th domain.
Each attribute value is assumed to be obtained by
integrating the GP $\bm{f}_v(\bm{x})$
over the corresponding support;
$\bm{y}_v$ is then generated from a Gaussian distribution\footnote{%
\label{fn:intg}We assume that the integral appearing in~\eqref{eq:o_model}
is well-defined. Sample paths of a GP are in general not integrable
without additional assumptions; the conditions under which
the integral is well-defined are discussed in Supplementary Material
of~\cite{tanaka:spatially}.},
\begin{flalign}
 \bm{y}_v \mid \bm{f}_v(\bm{x}), {\bf W}_v
 &\sim \mathcal{N}
 \left(
 \bm{y}_v \Bigm| \int_{\mathcal{X}_v} 
 {\bf A}_v(\bm{x})\bm{f}_v(\bm{x})\,d\bm{x},
 \bm{\Sigma}_v
 \right),
 \label{eq:o_model}
\end{flalign}
where ${\bf A}_v(\bm{x}):\mathcal{X}_v\to\mathbb{R}^{N_v\times|\mathcal{S}|}$
is represented by
\begin{flalign}
 {\bf A}_v(\bm{x})
 &= {\rm diag}\left(
 \bm{a}_{v1}(\bm{x}),\ldots,\bm{a}_{v|\mathcal{S}|}(\bm{x})
 \right),
\end{flalign}
in which
$\bm{a}_{vs}(\bm{x}) = 
(a_{vs1}(\bm{x}),\ldots,a_{vs|\mathcal{P}_{vs}|}(\bm{x}))^\top$,
whose entry $a_{vsn}(\bm{x})$ is a weight function
for aggregation over support $\mathcal{R}_{vsn}$.
This modeling does not depend on the particular choice of
$\{a_{vsn}(\bm{x})\}$ as long as they are integrable.
If one takes, for support $\mathcal{R}_{vsn}$,
\begin{flalign}
 a_{vsn}(\bm{x})
 &=\frac{{\mathbbm 1}(\bm{x}\in\mathcal{R}_{vsn})}
 {\int_{\mathcal{X}_v} {\mathbbm 1}(\bm{x}'\in\mathcal{R}_{vsn})
 \,d\bm{x}'},
 \label{eq:average_model}
\end{flalign}
where ${\mathbbm 1}(\bullet)$ is the indicator function;
${\mathbbm 1}(Z) =1$ if $Z$ is true and ${\mathbbm 1}(Z) =0$ otherwise,
then $y_{vsn}$ is the average of $f_{vs}(\bm{x})$
over $\mathcal{R}_{vsn}$.
One may also consider other aggregation processes
to suit the property of the attribute values,
including simple summation and population-weighted averaging
over $\mathcal{R}_{vsn}$.
$\bm{\Sigma}_v = {\rm diag}(\sigma_{v1}^2 {\bf I},\ldots,
\sigma_{v|\mathcal{S}|}^2 {\bf I})$ in (\ref{eq:o_model}) is
an $N_v\times N_v$ block diagonal matrix,
where $\sigma_{vs}^2$ is the noise variance
for the $s$-th attribute in the $v$-th domain,
where ${\bf I}$ is the identity matrix.

{\bf On domains.}
We here summarize our assumptions of A-MoGP in modeling multiple domains,
in order to avoid possible misunderstandings. 
The field $\bm{f}_v(\bm{x})$~\eqref{eq:product_wg} for attribute values
is assumed to be determined by
the product of the latent field $\bm{g}_v(\bm{x})$
and the weight matrix ${\bf W}_v$.
It should be noted that, for each domain $v\in\mathcal{V}$, 
$\bm{g}_v(\bm{x})$ and ${\bf W}_v$ in~\eqref{eq:product_wg}
are assumed to be realizations from the prior distributions
of~\eqref{eq:latent_GP} and~\eqref{eq:prior_w}, respectively, 
so that a single realization of $\bm{g}_v(\bm{x})$ on $\mathcal{X}_v$,
as well as a single realization of ${\bf W}_v$, 
is shared over the whole domain $\mathcal{X}_v$
to define the field $\bm{f}_v(\bm{x})$
on $\mathcal{X}_v$ via~\eqref{eq:product_wg}. 
Another assumption is that
$\{(\bm{g}_1(\bm{x}),{\bf W}_1),(\bm{g}_2(\bm{x}),{\bf W}_2),
\ldots,(\bm{g}_{|\mathcal{V}|}(\bm{x}),{\bf W}_{|\mathcal{V}|})\}$
are assumed independent. 
Due to this assumption we do not have to consider
cross-domain covariance ${\bf \Gamma}(\bm{x},\bm{x}')$
with $\bm{x}\in\mathcal{X}_v,\bm{x}'\in\mathcal{X}_{v'},v\not=v'$.
Meanwhile, we assume that the prior distributions for
$(\bm{g}_v(\bm{x}),{\bf W}_v)$ are shared across domains,
which allows domain transfer.
In the case where the set of attributes is different across domains,
we use the same priors for the same kind of attributes in different domains.

\section{Inference}
\label{sec:inference}
Our aim is to obtain the predictive distribution
on the basis of variational Bayesian inference procedures.
The model parameters,
$\{{\bf W}_v\mid v\in\mathcal{V}\}$,
$\{\bm{\Sigma}_v\mid v\in\mathcal{V}\}$,
$\bm{\Gamma}(\bm{x},\bm{x}')$,
$\{\bar{w}_{sl}\mid s\in\mathcal{S}; l\in\mathcal{L}\}$,
and $\{\eta_{sl}^2\mid s\in\mathcal{S}; l\in\mathcal{L}\}$,
are estimated by maximizing
the evidence lower bound (ELBO),
in which GPs are analytically integrated out.
We adopt the reparameterization trick~\cite{kingma:auto}
to estimate variational parameters
via gradient-based optimization methods.
One can subsequently obtain the predictive distribution
using the estimated parameters.
The inference procedure is shown in Algorithm~\ref{alg1}.

\begin{algorithm}[!t]
 \caption{Inference procedure for A-MoGP.}
 \label{alg1}
 \begin{algorithmic}[1]
  \Require{Aggregate datasets $\mathcal{D}$, the numbers of Monte-Carlo samples $T_{\rm e}$, $T_{\rm p}$}
  \Ensure{Predictive distribution $p(\bm{f}_v^* \mid \bm{x}, \mathcal{D})$}
  \State Initialize
  $\{\bm{\Sigma}_v\}$,
  $\bm{\Gamma}(\bm{x},\bm{x}')$,
  $\{\bar{w}_{sl}\}$,
  $\{\eta_{sl}^2\}$,
  $\{\bar{w}_{sl}'\}$,
  $\{\eta_{sl}'^2\}$
  \State /* Parameter learning */
  \Repeat
  \For{$t=1,\ldots,T_{\rm e}$}
  \ForAll{$v,s,l$}
  \State $\hat{w}_{vsl}^{(t)} \leftarrow
  \bar{w}_{vsl}'+\epsilon\cdot\sqrt{\eta_{vsl}'^2}$,
  $\;$where$\;$ $\epsilon\sim\mathcal{N}(0,1)$
  \EndFor
  \EndFor
  \State Update $\{\bm{\Sigma}_v\}$,
  $\bm{\Gamma}(\bm{x},\bm{x}')$,
  $\{\bar{w}_{sl}\}$,
  $\{\eta_{sl}^2\}$,
  $\{\bar{w}_{sl}'\}$,
  $\{\eta_{sl}'^2\}$
  by maximizing the ELBO~\eqref{eq:elbo}
  \Until{Convergence}
  \State /* Prediction */
  \For{$t=1,\ldots,T_{\rm p}$}
  \ForAll{$v,s,l$}
  \State $\hat{w}_{vsl}^{(t)} \leftarrow w_{vsl} \sim 
  \mathcal{N}\left(w_{vsl} \mid \bar{w}_{vsl}',\,\eta_{vsl}'
  \right)$~\eqref{eq:variational_dist}
  \EndFor
  \EndFor
  \State Construct the predictive distribution
  $p(\bm{f}_v^* \mid \bm{x}, \mathcal{D})$~\eqref{eq:mc_post}
  using the estimated parameters
 \end{algorithmic}
\end{algorithm}

{\bf ELBO}.
Given the aggregate datasets in multiple domains $\mathcal{D}$,
the marginal likelihood (i.e., evidence) of $\{\bm{y}_v\}$ is given by
\begin{flalign}
 p(\{\bm{y}_v\})
 &= \prod_{v\in\mathcal{V}}
 \int p(\bm{y}_v \mid {\bf W}_v)
 p({\bf W}_v)\,d{\bf W}_v.
 \label{eq:marginal}
\end{flalign}
Here, the likelihood of $\bm{y}_v$ is
\begin{flalign}
 p(\bm{y}_v\mid{\bf W}_v)
 &= \mathcal{N}
 \left(
 \bm{y}_v \mid \bm{0}, {\bf C}_v
 \right),
 \label{eq:likelihood}
\end{flalign}
where we analytically integrate out the GP prior $\bm{f}_v(\bm{x})$,
and where ${\bf C}_v$ is an $N_v\times N_v$ covariance matrix represented by
\begin{flalign}
 {\bf C}_v
 &= \iint_{\mathcal{X}_v\times\mathcal{X}_v}
 {\bf A}_v(\bm{x}) {\bf K}_v(\bm{x},\bm{x}') {\bf A}_v(\bm{x}')^\top
 \,d\bm{x}\,d\bm{x}' + \bm{\Sigma}_v.
 \label{eq:Cv}
\end{flalign}
It is an $|\mathcal{S}| \times |\mathcal{S}|$ block matrix
whose $(s,s')$-th block ${\bf C}_{vss'}$ is 
a $|\mathcal{P}_{vs}|\times|\mathcal{P}_{vs'}|$ matrix represented by
\begin{flalign}
 {\bf C}_{vss'}
 &=\iint_{\mathcal{X}_v\times\mathcal{X}_v}
 k_{vss'}(\bm{x},\bm{x}') 
 \bm{a}_{vs}(\bm{x})
 \bm{a}_{vs'}(\bm{x}')^\top
 \,d\bm{x}\,d\bm{x}'\nonumber \\
 &\quad+ \delta_{s,s'} \sigma_{vs}^2 {\bf I},
\end{flalign}
where $\delta_{\bullet,\bullet}$ represents Kronecker's delta;
$\delta_{Z,Z'} = 1$ if $Z=Z'$ and $\delta_{Z,Z'} = 0$ otherwise.
The closed-form expression~\eqref{eq:Cv} for ${\bf C}_v$ is not trivial
for the following reason:
One needs to consider averaging with respect to
the infinite-dimensional Gaussian $\bm{f}_v(\bm{x})$
since the attribute values are obtained
by aggregating the function values
evaluated at an infinite number of points
within the corresponding supports.
Our works~\cite{tanaka:spatially,tanaka:probabilistic} have shown that
the integral with respect to $\bm{f}_v(\bm{x})$
under the integrability conditions mentioned in footnote~\ref{fn:intg} 
can be analytically performed
by the following procedures:
Consider the Riemann sums to define the integral of $\bm{f}_v(\bm{x})$
using a regular grid covering $\mathcal{X}_v$
with the grid cell volume $\Delta$;
integrate out the GP prior in a similar way to the vanilla GP;
take the limit $\Delta\rightarrow0$,
obtaining the likelihood~\eqref{eq:likelihood}.
Details appear in Supplementary Material of~\cite{tanaka:spatially}.
Equation~\eqref{eq:Cv} takes the form of the double integral
of the covariance function
${\bf K}_v(\bm{x},\bm{x}')$
over the respective pairs of supports,
which conceptually corresponds to
the aggregation of the covariance function values
that are calculated at the infinite pairs of points
in the corresponding supports;
this allows one to evaluate the support-to-support covariances
taking into account support size and shape.
How to evaluate the integral of the covariance function in~\eqref{eq:Cv} is described
at the end of this section.
In general, it is difficult to evaluate
the logarithm of the evidence~\eqref{eq:marginal},
so that we consider what is called the evidence lower bound (ELBO), defined as 
\begin{flalign}
 &\ln p(\{\bm{y}_v\})
 \nonumber\\
 &\;\;\geq \sum_{v\in\mathcal{V}}\left\{
 \mathbb{E}_{q({\bf W}_v)}\left[
 \ln p(\bm{y}_v \mid {\bf W}_v)
 \right] - {\rm KL}\left[
 q({\bf W}_v)\,\|\, p({\bf W}_v)
 \right]
 \right\},
 \label{eq:elbo}
\end{flalign}
where $\mathbb{E}[\bullet]$ is the expectation,
${\rm KL}[\bullet\|\bullet]$ is the Kullback-Leibler (KL) divergence,
and $q({\bf W}_v)$ is a variational distribution
that approximates the true posterior $p({\bf W}_v \mid \{\bm{y}_v\})$.
On the basis of the mean-field approximation~\cite{bishop:pattern},
we take, for each domain, the variational distribution given by
a factorized Gaussian distribution,
\begin{flalign}
 q({\bf W}_v)
 &= \prod_{s\in\mathcal{S}} \prod_{l\in\mathcal{L}}
 \mathcal{N}\left(
 w_{vsl} \mid \bar{w}_{vsl}',\,\eta_{vsl}'^2
 \right),
 \label{eq:variational_dist}
\end{flalign}
where $\bar{w}_{vsl}'$ and $\eta_{vsl}'^2$ are
the element-wise variational parameters.
It is, unfortunately, difficult to directly obtain the ELBO
as the expectation in~\eqref{eq:elbo} is not tractable;
we compute it using Monte-Carlo approximation as follows:
\begin{flalign}
 \mathbb{E}_{q({\bf W}_v)}\left[
 \ln p(\bm{y}_v \mid {\bf W}_v)
 \right]
 &\approx \frac{1}{T_{\rm e}}\sum_{t=1}^{T_{\rm e}}
 \ln p(\bm{y}_v \mid \hat{{\bf W}}_v^{(t)}),
 \label{eq:mc_elbo}
\end{flalign}
where $\hat{{\bf W}}_v^{(t)}\sim q({\bf W}_v)$
and $T_{\rm e}$ is the number of Monte-Carlo samples
for evaluating the ELBO.
We use the reparameterization trick~\cite{kingma:auto},
so that the $t$-th sample $\hat{w}_{vsl}^{(t)}$ of $\hat{w}_{vsl}$
is given by
\begin{flalign}
 \hat{w}_{vsl}^{(t)}
 &= \bar{w}_{vsl}'+\epsilon\cdot\sqrt{\eta_{vsl}'^2},
\end{flalign}
where $\epsilon\sim\mathcal{N}(0,1)$.
This technique allows us to use gradient-based optimization methods
for estimating the variational parameters.
Accordingly, the model parameters and the variational parameters
can be obtained by maximizing the ELBO~\eqref{eq:elbo}.

{\bf Predictive distribution}.
Let $\bm{f}^*_v$ denote the predictive value
of $\bm{f}(\bm{x})$ at $\bm{x}$.
We first describe the posterior distribution
for $\bm{f}^*_v$
conditional on ${\bf W}_v$,
obtained in the closed form as
\begin{flalign}
 p(\bm{f}_v^*\mid\bm{x},{\bf W}_v,\mathcal{D})
 &= \mathcal{N}\left(
 \bm{f}_v^*\mid
 \bm{m}_v^*(\bm{x}),
 {\bf K}_v^*(\bm{x},\bm{x}')
 \right),
 \label{eq:conditional_post}
\end{flalign}
where $\bm{m}_v^*(\bm{x}): \mathcal{X}_v\to\mathbb{R}^{|\mathcal{S}|}$
and ${\bf K}_v^*(\bm{x}, \bm{x}'):
\mathcal{X}_v\times\mathcal{X}_v\to\mathbb{R}^{|\mathcal{S}|\times|\mathcal{S}|}$ are
the mean function and the covariance function, respectively,
for the conditional posterior.
Defining 
${\bf H}_v(\bm{x}):\mathcal{X}_v\to\mathbb{R}^{N_v\times|\mathcal{S}|}$ as
\begin{flalign}
 {\bf H}_v(\bm{x})
 &= \int_{\mathcal{X}_v}
 {\bf A}_v(\bm{x}') {\bf K}_v(\bm{x}', \bm{x})\,d\bm{x}',
 \label{eq:H}
\end{flalign}
which consists of the point-to-support covariances,
the mean function $\bm{m}_v^*(\bm{x})$
and the covariance function ${\bf K}_v^*(\bm{x}, \bm{x}')$
are given by
\begin{flalign}
 \bm{m}_v^*(\bm{x})
 &= {\bf H}_v(\bm{x})^\top {\bf C}_v^{-1} \bm{y}_v,
 \label{eq:post_mean}\\
 {\bf K}_v^*(\bm{x}, \bm{x}')
 &= {\bf K}_v(\bm{x}, \bm{x}') 
 - {\bf H}_v(\bm{x})^\top {\bf C}_v^{-1} {\bf H}_v(\bm{x}'),
 \label{eq:post_var}
\end{flalign}
respectively.
Derivation of the conditional posterior~\eqref{eq:conditional_post}
is similar to that of the likelihood~\eqref{eq:likelihood}.
According to variational Bayesian inference procedures,
the predictive distribution of $\bm{f}^*_v$ is given by
the expectation of $\bm{f}^*_v$
with respect to the variational distribution of ${\bf W}_v$;
however, it is intractable;
we thus compute it using Monte-Carlo approximation as follows:
\begin{flalign}
 p(\bm{f}_v^* \mid \bm{x}, \mathcal{D})
 &= \int p(\bm{f}_v^* \mid \bm{x}, {\bf W}_v,\mathcal{D}) q({\bf W}_v) \,d{\bf W}_v
 \nonumber\\
 &\approx \frac{1}{T_{\rm p}}\sum_{t=1}^{T_{\rm p}}
 p(\bm{f}_v^*\mid\bm{x},\hat{{\bf W}}_v^{(t)},\mathcal{D})
 \label{eq:mc_post}
\end{flalign}
where $\hat{{\bf W}}_v^{(t)}\sim q({\bf W}_v)$,
and where $T_{\rm p}$ is the number of Monte-Carlo samples
to approximate the predictive distribution.
One can observe that the approximate distribution~\eqref{eq:mc_post}
is the form of a multivariate Gaussian mixture.
Then, one can obtain the mean function $\hat{\bm{m}}_v^*(\bm{x})$ and
the covariance function $\hat{{\bf K}}_v^*(\bm{x},\bm{x}')$
for the approximate predictive distribution,
represented by
\begin{flalign}
 \hat{\bm{m}}_v^*(\bm{x})
 &=\frac{1}{T_{\rm p}}
 \sum_{t=1}^{T_{\rm p}}\bm{m}_v^{*(t)}(\bm{x}),
 \label{eq:VB_post_mean}\\
 \hat{{\bf K}}_v^*(\bm{x},\bm{x}')
 &=\frac{1}{T_{\rm p}} \sum_{t=1}^{T_{\rm p}} 
 \left(
 {\bf K}_v^{*(t)}(\bm{x},\bm{x}')
 + \bm{m}_v^{*(t)}(\bm{x}) \bm{m}_v^{*(t)}(\bm{x}')^\top
 \right)\nonumber\\
 &\quad -\hat{\bm{m}}_v^*(\bm{x})\hat{\bm{m}}_v^*(\bm{x}')^\top
 \label{eq:VB_post_cov}
\end{flalign}
Here, $\bm{m}_v^{*(t)}(\bm{x})$ and
${\bf K}_v^{*(t)}(\bm{x},\bm{x}')$ are
the mean function~\eqref{eq:post_mean}
and the covariance function~\eqref{eq:post_var}, respectively,
that are evaluated using the $t$-th sample
$\hat{{\bf W}}_v^{(t)}$.

{\bf Integral of the covariance function}.
In~\eqref{eq:Cv} and~\eqref{eq:H},
we need the integrals of the covariance function
${\bf K}_v(\bm{x},\bm{x}')$ over the domain $\mathcal{X}_v$.
If the dimension $D$ of $\mathcal{X}_v$ is one,
it can be obtained in a closed form 
as long as the covariance function is analytically integrable.
One can find an example using the squared-exponential kernel
as the covariance function in Section~2 of~\cite{smith:gaussian},
which we adopt in the following experiments.
If $D\geq2$, on the other hand, 
this integral generally requires a numerical approximation.
In the implementation, we approximate these integrals by using
sufficiently fine-grained $D$-dimensional square grid cells.
We divide domain $\mathcal{X}_v$ into square grid cells,
and take $\mathcal{G}_{vsn}$ to be the set of grid points
that are contained in support $\mathcal{R}_{vsn}$.
Consider the approximation of the integral
in the covariance matrix~\eqref{eq:Cv}.
The $(n,n')$-entry $C_{vss'}(n,n')$
of ${\bf C}_{vss'}$ is approximated as follows:
\begin{flalign}
 &C_{vss'}(n,n')\nonumber\\
 &\;\;=\iint_{\mathcal{X}_v\times\mathcal{X}_v} 
 k_{vss'}(\bm{x}, \bm{x}') 
 a_{vsn}(\bm{x}) a_{vs'n}(\bm{x}')\,d\bm{x}\,d\bm{x}'
 + \delta_{s,s'} \sigma^2_{vs} \nonumber\\
 &\;\;\approx \frac{1}{|\mathcal{G}_{vsn}|}
 \frac{1}{|\mathcal{G}_{vs'n'}|} \sum_{i\in\mathcal{G}_{vsn}} 
 \sum_{j\in\mathcal{G}_{vs'n'}}
 k_{vss'}(i,j) + \delta_{s,s'} \sigma^2_{vs},
 \label{eq:approx}
\end{flalign}
where we use the formulation of
the support-average-observation model~\eqref{eq:average_model}.
The integrals in~\eqref{eq:H} can be approximated in a similar way.
Letting $|{\mathcal G}_v|$ denote the number of all grid points
in the $v$-th domain,
the computational complexity of ${\bf C}_{vss'}$~\eqref{eq:Cv}
is $O(|\mathcal{G}_v|^2)$.
Assuming the constant weight $a_{vsn}(\bm{x}) = a_{vsn}$
(e.g., support-average),
the computational complexity can be reduced to
$O(|\mathcal{P}_{vs}| |\mathcal{P}_{vs'}| |\mathcal{E}|)$,
where $|\mathcal{E}|$ is the cardinality of the set of
distinct distance values between grid points.
Here, we use the property that $k_{vss'}(i,j)$ in~\eqref{eq:approx}
depends only on the distance between $i$ and $j$,
which is helpful for saving the computation time
and the memory requirement.
The average computation time for training was 738.7 seconds
for the areal datasets from New York City and Chicago;
the experiments were conducted
on an NVIDIA Titan RTX GPU.

\section{Experiments}
\label{sec:experiments}
\subsection{Datasets}
\label{sec:data}
We evaluated the proposed model, A-MoGP,
using real-world datasets defined
in one- and two-dimensional domains.
The one-dimensional domain is a temporal domain,
and the attribute values are obtained
by aggregating time series on time bins.
The two-dimensional domain is a spatial domain;
the attributes are aggregated on spatial regions.
Details of the datasets are described in subsequent paragraphs.
In the experiments, the attribute values were centered and normalized
so that each attribute in each domain has zero mean and unit variance.

{\bf Time series of air pollutants.}
This dataset includes hourly air pollutant concentrations
from multiple air quality monitoring stations
in Beijing~\cite{zhang:cautionary};
the observation period is from March 1st, 2013, to February 28th, 2017.
For evaluation, we used three pollutants, $\mathrm{NO}_2,\mathrm{CO},\mathrm{O}_3$, which are denoted in the following by NO2, CO, and O3, respectively, 
from three monitoring stations, Changping, Aotizhongxin, and Dingling.
The observation period was divided equally into three disjoint parts,
each of which was regarded as the {\em domain} for each monitoring station.
In other words, in the time series datasets
we used in the following experiments,
there are three domains, that is, three observation periods
for respective monitoring stations.
We explored the task of refining the time series aggregated
on the coarse-grained time bins
by using the aggregated time series from multiple monitoring stations.
Here, the {\em support} in the time series datasets
corresponds to each of the time bins.
To evaluate the performance in predicting the fine-grained attribute,
we first picked up one target dataset
and used its coarser version for learning model parameters;
then, we predicted the fine-grained target attribute via the learned model.
We created the coarse- and fine-grained target attributes
by aggregating the original data on time bins of different sizes. 
More concretely, we prepared three kinds of coarse-grained time bins
to evaluate performance when changing bin sizes as one, two, and four months.
The bin size of the fine-grained target attribute was set to one week.

\begin{table}[!t]
 \caption{Real-world areal datasets.}
 \label{tb:areal_data}
 \begin{center}
  \subtable[New York City]{
  \begin{tabular}{l l r r} \toprule
   Attribute &Partition &\#regions &Time range \\ \midrule
   Poverty rate &Community &59 &2009 -- 2013 \\
   Unemployment rate &Community &59 &2009 -- 2013 \\
   PM2.5 &UHF42 &42 &2009 -- 2010 \\
   Mean commute &Community &59 &2009 -- 2013 \\
   Recycle diversion rate &Community &59 &2009 -- 2013 \\
   Population density &Community &59 &2009 -- 2013 \\
   Crime rate &Police precinct &77 &2010 -- 2016 \\
   \bottomrule
  \end{tabular}
  }
  \\
  \subtable[Chicago]{
  \begin{tabular}{l l r r} \toprule
   Attribute &Partition &\#regions &Time range \\ \midrule
   Poverty rate &Community &77 &2008 -- 2012 \\
   Unemployment rate &Community &77 &2008 -- 2012 \\
   Crime rate &Police precinct &25 &2012 \\
   \bottomrule
  \end{tabular}
  }
 \end{center}
\end{table}

{\bf Areal data in cites.}
We conducted the experiments
using seven and three real-world areal datasets
from New York City~\cite{NYCdata}
and Chicago~\cite{CHIdata},
respectively.
There are two {\em domains} in the areal datasets,
where the domain corresponds to the total region of each city.
The areal dataset is associated with
one of the predefined geographical partitions with various granularities:
UHF42 (42), Community (59), and Police precinct (77) in New York City;
Police precinct (25), and Community (77) in Chicago,
where each number in parentheses denotes
the number of spatial regions in the corresponding partition.
In the areal datasets,
the spatial region corresponds to
the {\em support} for data aggregation.
Details about the real-world areal datasets are shown
in Table~\ref{tb:areal_data}.
These datasets were gathered once a year
at the time ranges shown in Table~\ref{tb:areal_data};
thus, we used the datasets by yearly averaging their attribute values.
The evaluation procedure is similar to the case
of the time series datasets (described in the preceding paragraph).
For evaluation, we created the coarse-grained target attributes
by aggregating the original data
on the coarse-grained partition:
Borough (5) in New York City and Side (9) in Chicago.

{\bf Evaluation metric.}
Let $v'\in\mathcal{V}$ and $s'\in\mathcal{S}_{v'}$ be the domain index and the attribute index we targeted, respectively.
The evaluation metric is the mean absolute percentage error (MAPE)
of the fine-grained target attribute values, represented by
\begin{flalign}
\frac{1}{|{\mathcal P}_{v's'}|} \sum_{n \in {\mathcal P}_{v's'}}
\left|\frac{y^{\rm true}_{v's'n} - y^\ast_{v's'n}}
{y^{\rm true}_{v's'n}}\right|,
\end{flalign}
where $y^{\rm true}_{v's'n}$ is the true value
associated with the $n$-th support
in the target fine-grained partition;
$y^\ast_{v's'n}$ is its predicted value,
obtained by integrating the $s'$-th posterior mean function
of the $v'$-th domain,
$\hat{\bm{m}}_{v'}^*(\bm{x})$~\eqref{eq:VB_post_mean},
over the corresponding target fine-grained support.

\subsection{Setup of A-MoGP}
\label{sec:setup}
In the experiments,
we used the squared-exponential kernel
as the covariance function for the latent GPs,
represented by
\begin{flalign}
\gamma_l(\bm{x}, \bm{x}')
= \alpha^2_l \exp \left(
- \frac{1}{2\beta^2_l}
\|\bm{x} - \bm{x}' \|^2
\right),
\end{flalign}
where $\alpha^2_l$ is a signal variance
that controls the magnitude of the covariance,
where $\beta_l$ is a scale parameter
that determines the covariances of data points,
and where $\|\cdot\|$ is the Euclidean norm.
Here, we set $\alpha^2_l = 1$ because the variance can already be
modeled by scaling the columns of ${\bf W}_v$ in~\eqref{eq:kernel}.
The model parameters were estimated by maximizing the ELBO~\eqref{eq:elbo}
using the Adam optimizer~\cite{kingma:adam}
with learning rate 0.001,
implemented in PyTorch~\cite{pytorch}.
As described in the last paragraph of Section~\ref{sec:inference},
we need to approximate the integral of kernel over the regions
in the areal data setting.
We divided the total region of each city
into sufficiently fine-grained square grid cells,
the size of which was 300~m $\times$ 300~m for both cities;
the resulting sets of grid points ${\mathcal G}_v$
for New York City and Chicago include
9,352 and 7,400 grid points, respectively.
The number $|\mathcal{L}|$ of the latent GPs was chosen from $\{1,\ldots,S\}$
via leave-one-out cross-validation~\cite{bishop:pattern},
where $S=\sum_{v\in\mathcal{V}}|\mathcal{S}_v|$.
We obtained the validation error using each held-out coarse-grained attribute value,
namely, we did not use the fine-grained target data in the validation process.
We set the numbers of Monte-Carlo samples as 
$T_{\rm e}=1$ and $T_{\rm p}=100$,
where the choice $T_{\rm e}=1$ amounts to
estimating the variational parameters via stochastic gradient ascent
of the ELBO with sample size 1. 

\begin{figure*}[tb]
 \centering
 \includegraphics[width=167mm, bb=0 0 1728 1692]{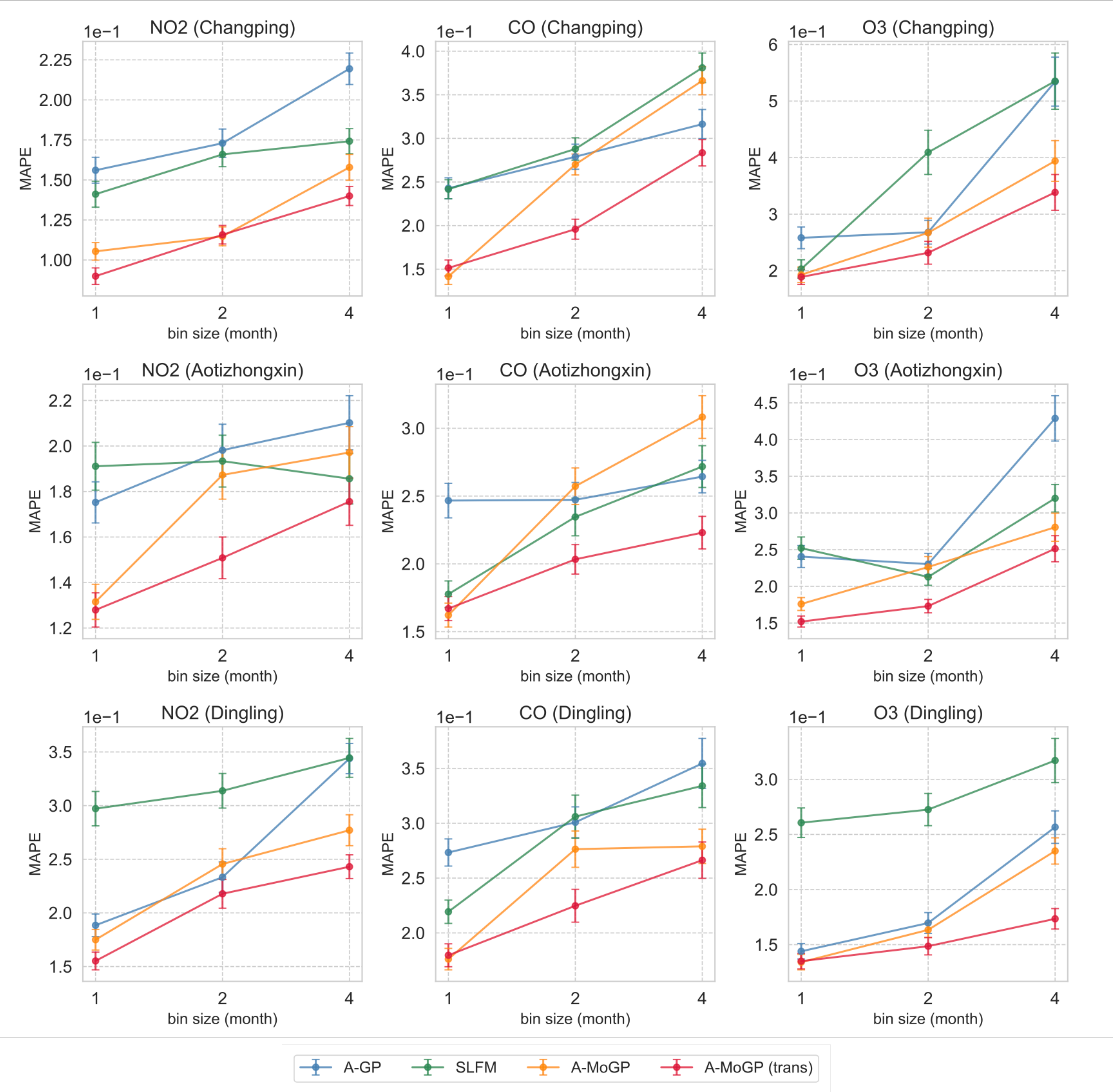}
 \caption{MAPE and standard errors for the prediction
   of fine-grained aggregated time-series datasets.
   Each row shows the results for each of the monitoring stations,
   and each column shows the results for each of the pollutants.}
 \label{fig:1d_MAPE}
\end{figure*}

\subsection{Baselines}
\label{sec:baselines}
We compared A-MoGP
with Gaussian process regression for binned data~\cite{smith:gaussian}.
This model is a special case of A-MoGP
when only target aggregate dataset in a single domain is available;
in this article,
we call it {\em aggregated Gaussian process (A-GP)}
with single-output.
Another baseline is one of the standard MoGP models,
semiparametric latent factor model (SLFM)~\cite{teh:semiparametric}.
A-MoGP is an extension of SLFM;
A-MoGP newly incorporates the observation model
with aggregation processes for handling aggregate data.
SLFM assumes that samples are observed at location points rather than supports;
we thus associate each attribute value
with the centroid of the support.
Additionally,
A-GP and SLFM do not have a mechanism
for knowledge transfer across domains.

\subsection{Results for time-series data}
\label{sec:1d_result}
\begin{figure*}[!t]
 \centering
 \includegraphics[width=180mm, bb=0 0 2016 564]{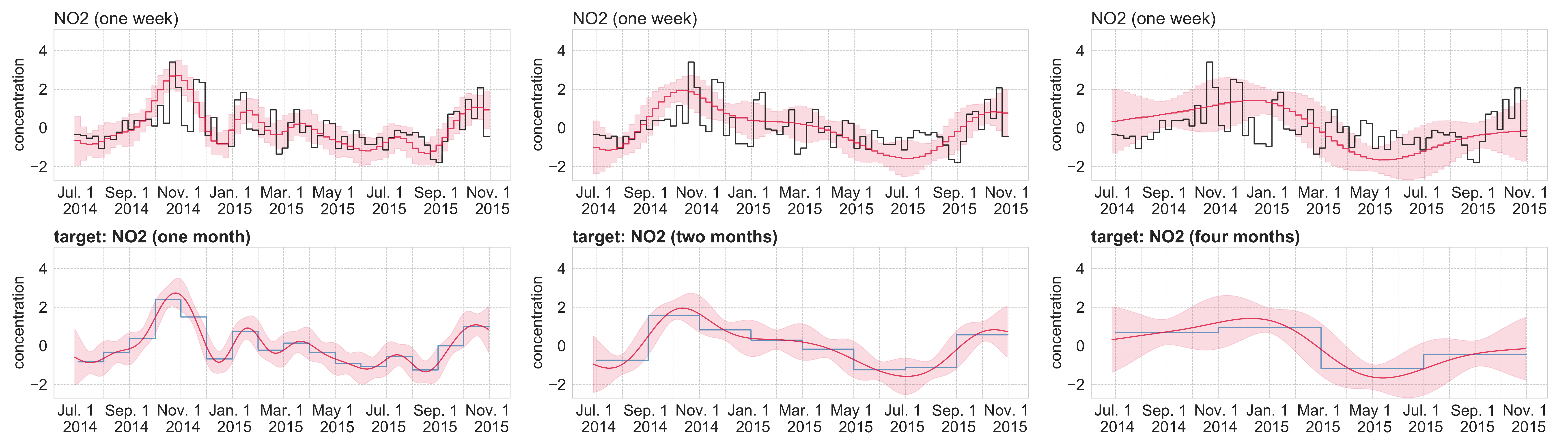}
 \caption{Prediction result of A-GP for the attributes in Aotizhongxin.
 In the first row,
 the black and red lines are the true and predicted values, respectively;
 the red shaded area denotes twice the standard deviation
 in prediction at each fine-grained bin.
 The blue and red lines in the second row are
 the training data and its prediction, respectively;
 the predictive variance is calculated on a continuous timeline.
 }
 \label{fig:vis_A-GP}
\end{figure*}
\begin{figure*}[!t]
 \vspace{.5cm}
 \centering
 \includegraphics[width=180mm, bb=0 0 2016 1152]{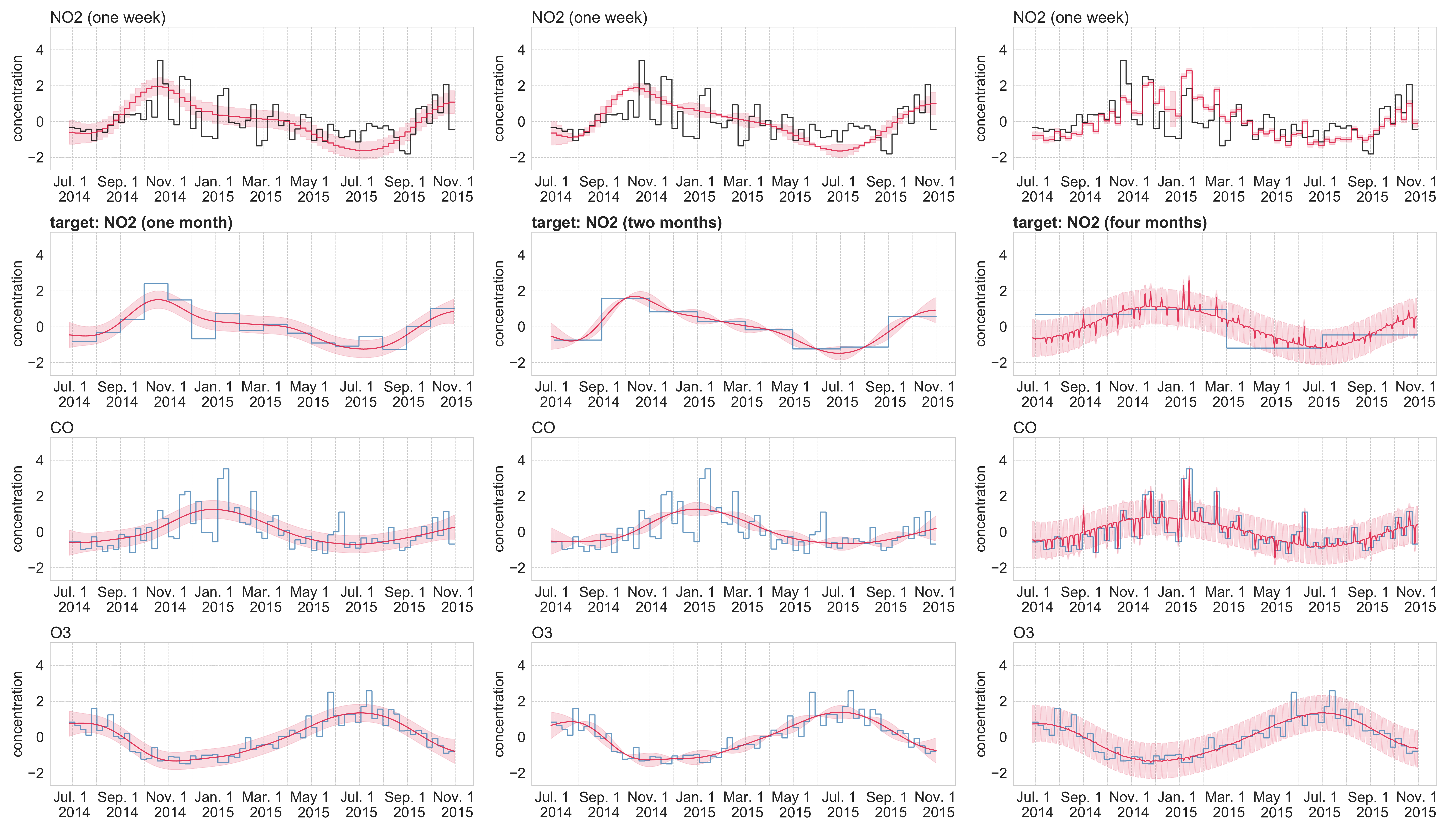}
 \caption{Prediction result of SLFM for the attributes in Aotizhongxin.
   Further figure details are the same as Figure~\ref{fig:vis_A-GP}.}
 \label{fig:vis_SLFM}
\end{figure*}
\begin{figure*}[!t]
 \vspace{.5cm}
 \centering
 \includegraphics[width=180mm, bb=0 0 2016 1152]{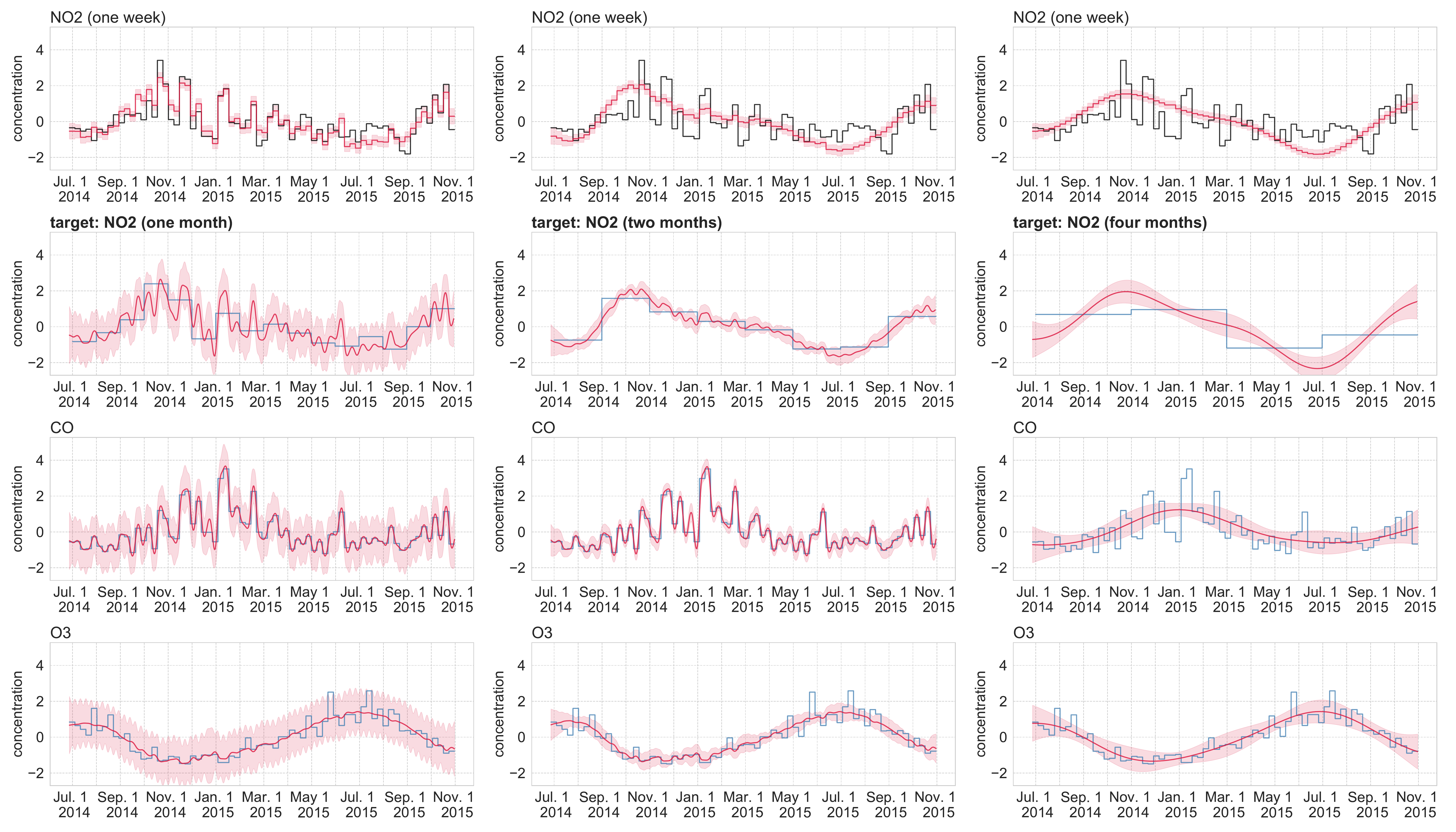}
 \caption{Prediction result of A-MoGP for the attributes in Aotizhongxin.
   Further figure details are the same as Figure~\ref{fig:vis_A-GP}.}
 \label{fig:vis_A-MoGP}
 \vspace{.5cm}
\end{figure*}
\begin{figure*}[!t]
 \centering
 \includegraphics[width=180mm, bb=0 0 2016 1152]{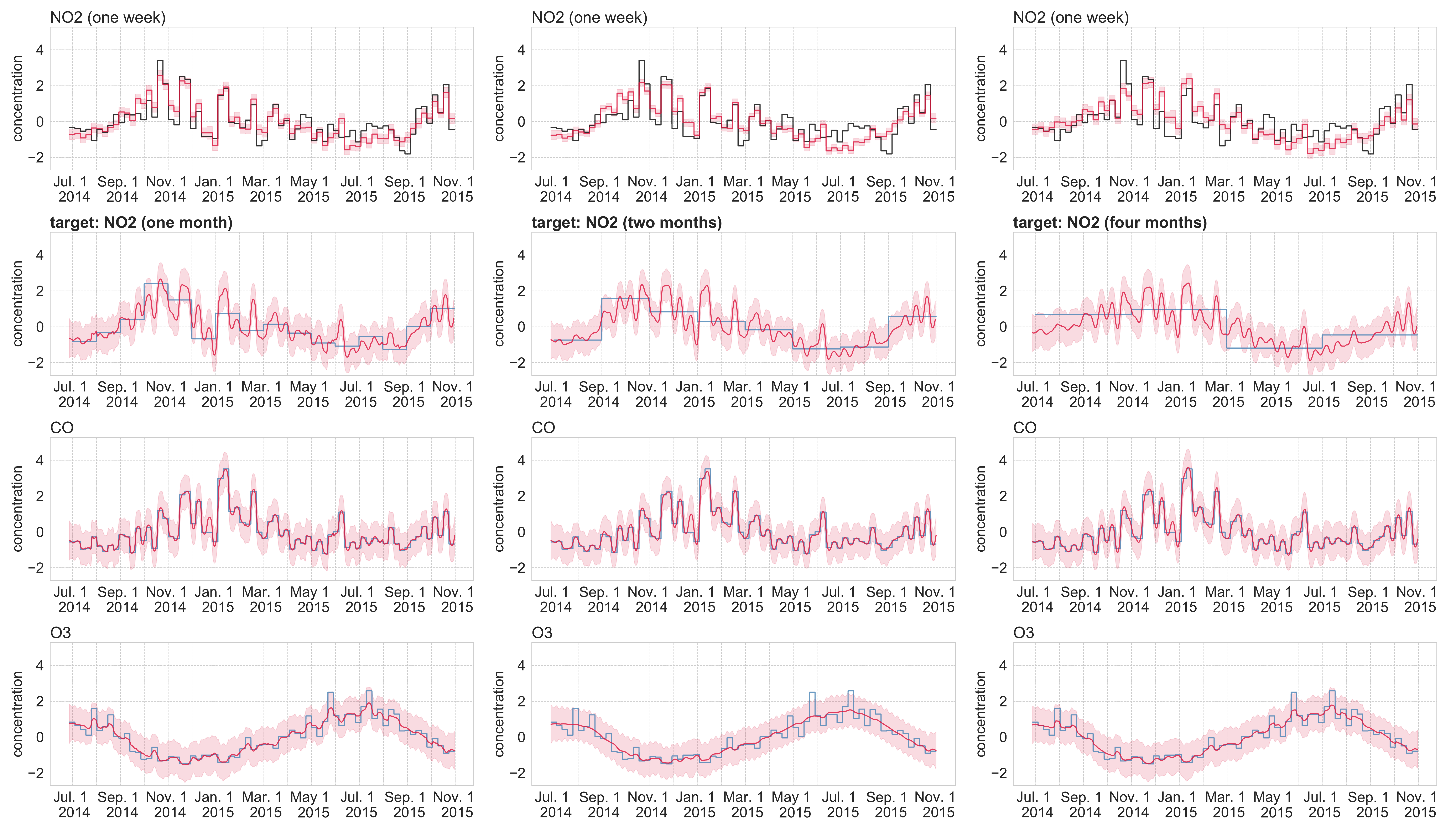}
 \caption{Prediction result of A-MoGP (trans) for the attributes in Aotizhongxin.
   Further figure details are the same as Figure~\ref{fig:vis_A-GP}.}
 \label{fig:vis_A-MoGP(trans)}
\end{figure*}
\begin{figure}[!t]
 \centering
 \subfigure[Relationships estimated by A-MoGP.]
 {\includegraphics[width=88mm, bb=0 0 2882 1004]{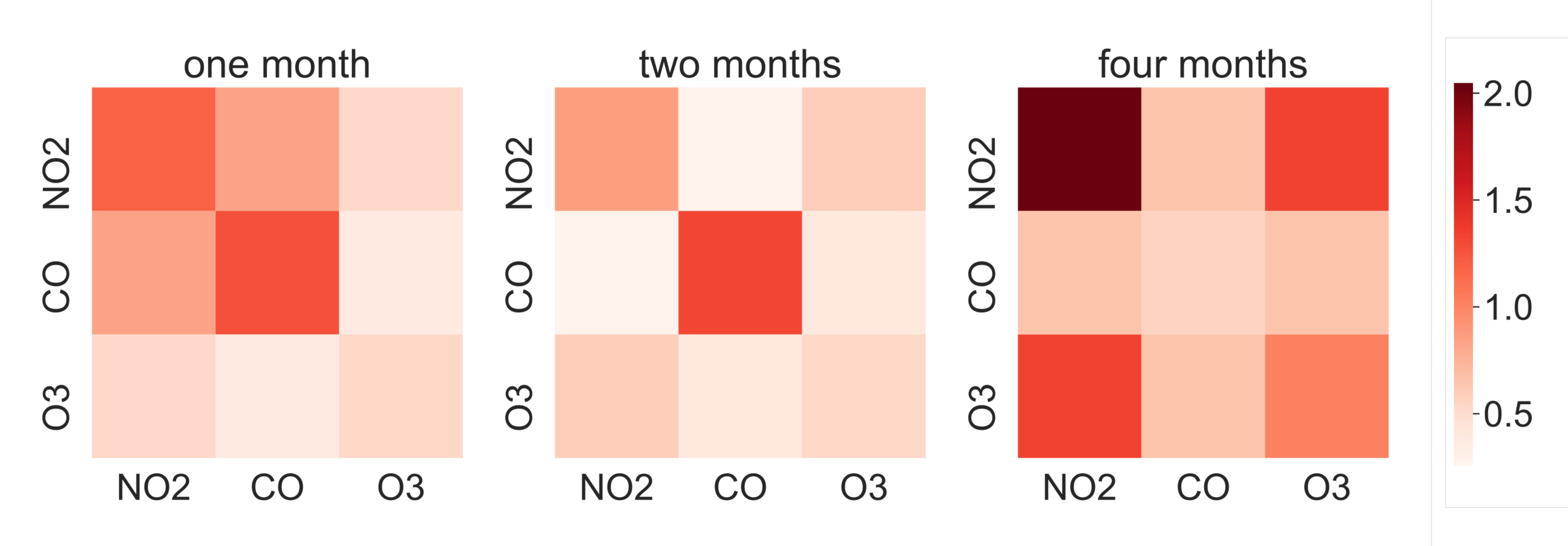}\label{}}
 \subfigure[Relationships estimated by A-MoGP (trans).]
 {\includegraphics[width=88mm, bb=0 0 2882 1004]{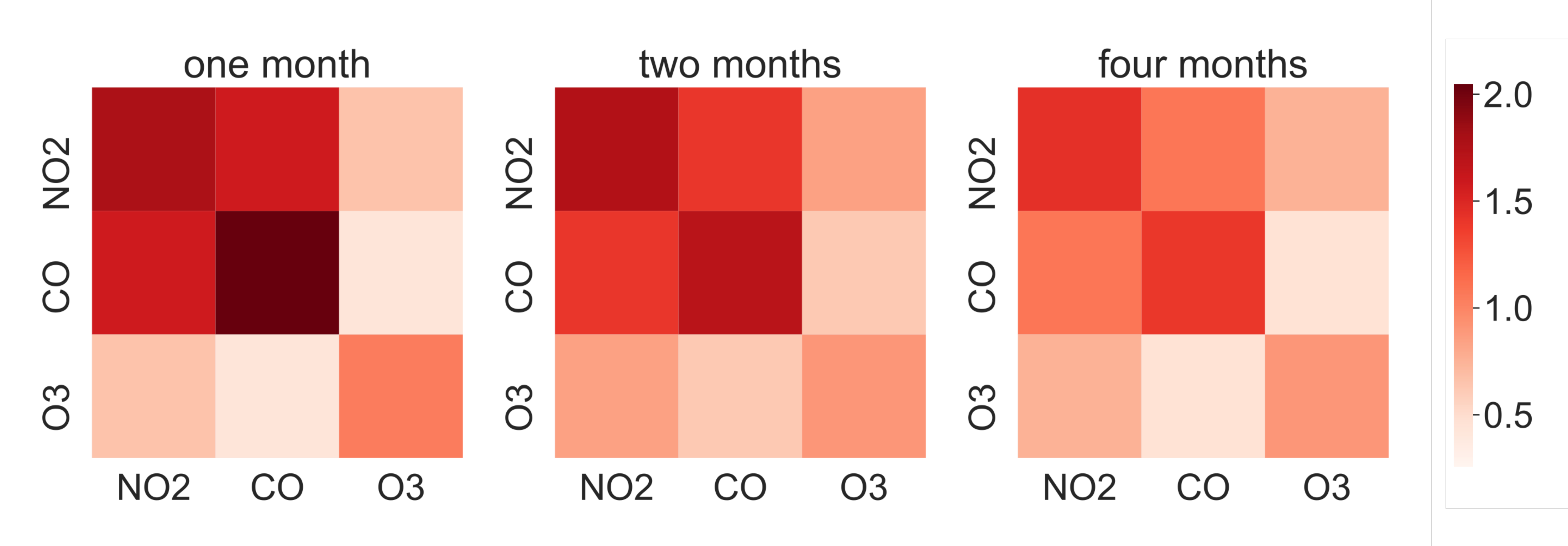}\label{}}
 \caption{Estimated relationships between attributes
 when using NO2 in Aotizhongxin as the target dataset.}
 \label{fig:relationship}
\end{figure}

In this section, we present the experimental results
for time-series data of air pollutants.
Figure~\ref{fig:1d_MAPE} shows MAPE and standard errors
for A-GP, SLFM, A-MoGP, and A-MoGP (trans).
Here, A-GP, SLFM, and A-MoGP were trained and tested
using the aggregate datasets from a single domain;
A-MoGP (trans) utilized all the datasets from three domains.
As expected, one observes that for all models,
MAPE increased with the bin size
for creating the coarse-grained target dataset.
In many cases, A-MoGP achieved better performance
than A-GP and SLFM.
This result shows that A-MoGP can effectively utilize
multiple aggregate datasets from a single domain.
Moreover, A-MoGP (trans) outperformed A-MoGP,
especially when the bin size was set to two and four months.
This result shows that
the refinement performance can be improved
by making use of aggregate datasets from other domains
even if the target dataset is coarser.

\begin{table*}[t!]
 \caption{MAPE and standard errors for the prediction of fine-grained 
 areal data in New York City and Chicago. 
 The numbers in parentheses denote the number $L$ of the latent GPs
 estimated by the leave-one-out cross-validation.}
 \label{tb:result_2d}
 \begin{center}
 \subtable[New York City]{\label{tb:}
  \begin{tabular}{l c c c c c} \toprule
   &A-GP &SLFM &A-MoGP &A-MoGP (trans) \\ \midrule
   Poverty rate &0.263 $\pm$ 0.032 (--) &0.184 $\pm$ 0.019 (2) &0.186 $\pm$ 0.018 (3) &{\bf 0.168 $\pm$ 0.018} (3)\\[1pt]
   Unemployment rate & 0.215 $\pm$ 0.023 (--) &0.172 $\pm$ 0.020 (6) &0.177 $\pm$ 0.021 (7) &{\bf 0.151 $\pm$ 0.020} (6)\\[1pt]
   PM2.5 &0.054 $\pm$ 0.007 (--) &{\bf 0.037 $\pm$ 0.005} (3) &{\bf 0.037 $\pm$ 0.006} (3) &0.039 $\pm$ 0.007 (4)\\[1pt]
   Mean commute &0.072 $\pm$ 0.009 (--) &0.072 $\pm$ 0.009 (6) &0.054 $\pm$ 0.006 (3) &{\bf 0.051 $\pm$ 0.006} (4)\\[1pt]
   Recycle diversion rate &0.282 $\pm$ 0.037 (--) &0.257 $\pm$ 0.029 (4) &0.180 $\pm$ 0.022 (2) &{\bf 0.166 $\pm$ 0.021} (4)\\[1pt]
   Population density &0.354 $\pm$ 0.039 (--) &0.386 $\pm$ 0.046 (4) &0.355 $\pm$ 0.040 (3) &{\bf 0.325 $\pm$ 0.052} (4)\\[1pt]
   Crime rate &0.601 $\pm$ 0.129 (--) &0.472 $\pm$ 0.081 (3) &0.427 $\pm$ 0.105 (2) &{\bf 0.426 $\pm$ 0.124} (8)\\[1pt]
   \bottomrule
  \end{tabular}}
  \\
 \subtable[Chicago]{\label{tb:}
  \begin{tabular}{l c c c c c} \toprule
   &A-GP &SLFM &A-MoGP &A-MoGP (trans) \\ \midrule
   Poverty rate &0.255 $\pm$ 0.031 (--) &0.309 $\pm$ 0.030 (2) &0.243 $\pm$ 0.031 (3) &{\bf 0.223 $\pm$ 0.021} (7) \\[1pt]
   Unemployment rate &0.414 $\pm$ 0.073 (--) & 0.356 $\pm$ 0.049 (2) &{\bf 0.240 $\pm$ 0.025} (3) &0.268 $\pm$ 0.037 (8) \\[1pt]
   \bottomrule
  \end{tabular}}
 \end{center}
\end{table*}
\stepcounter{figure}
\begin{figure*}[tb]
 \begin{center}
  \subfigure[Training data (coarse granularity)]
  {\includegraphics[width=50mm, bb=0 0 622.9125 548.2649703756]{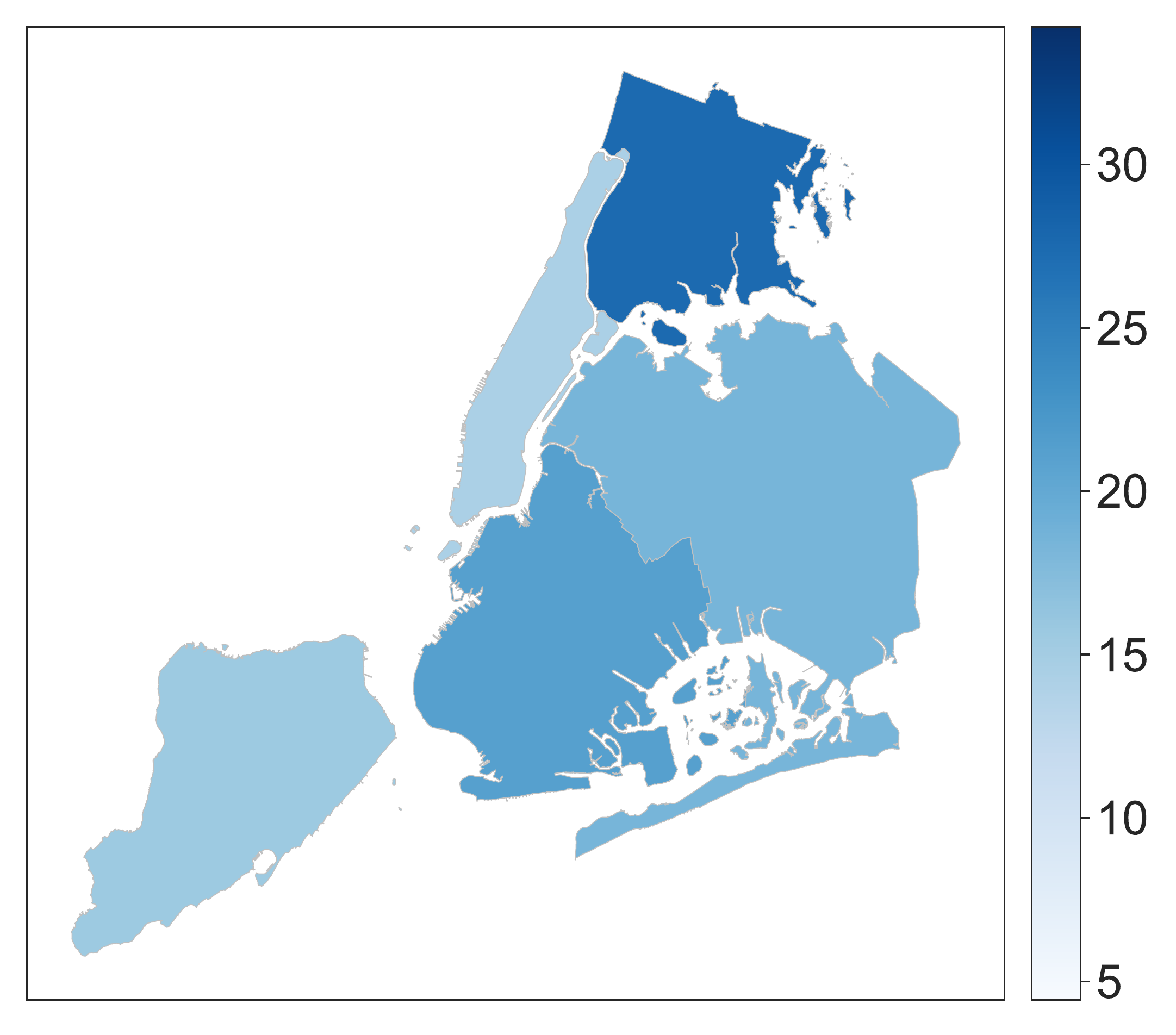}\label{fig:}}
  \hspace{.2cm}
  \subfigure[True data (fine granularity)]
  {\includegraphics[width=50mm, bb=0 0 622.9125 548.2649703756]{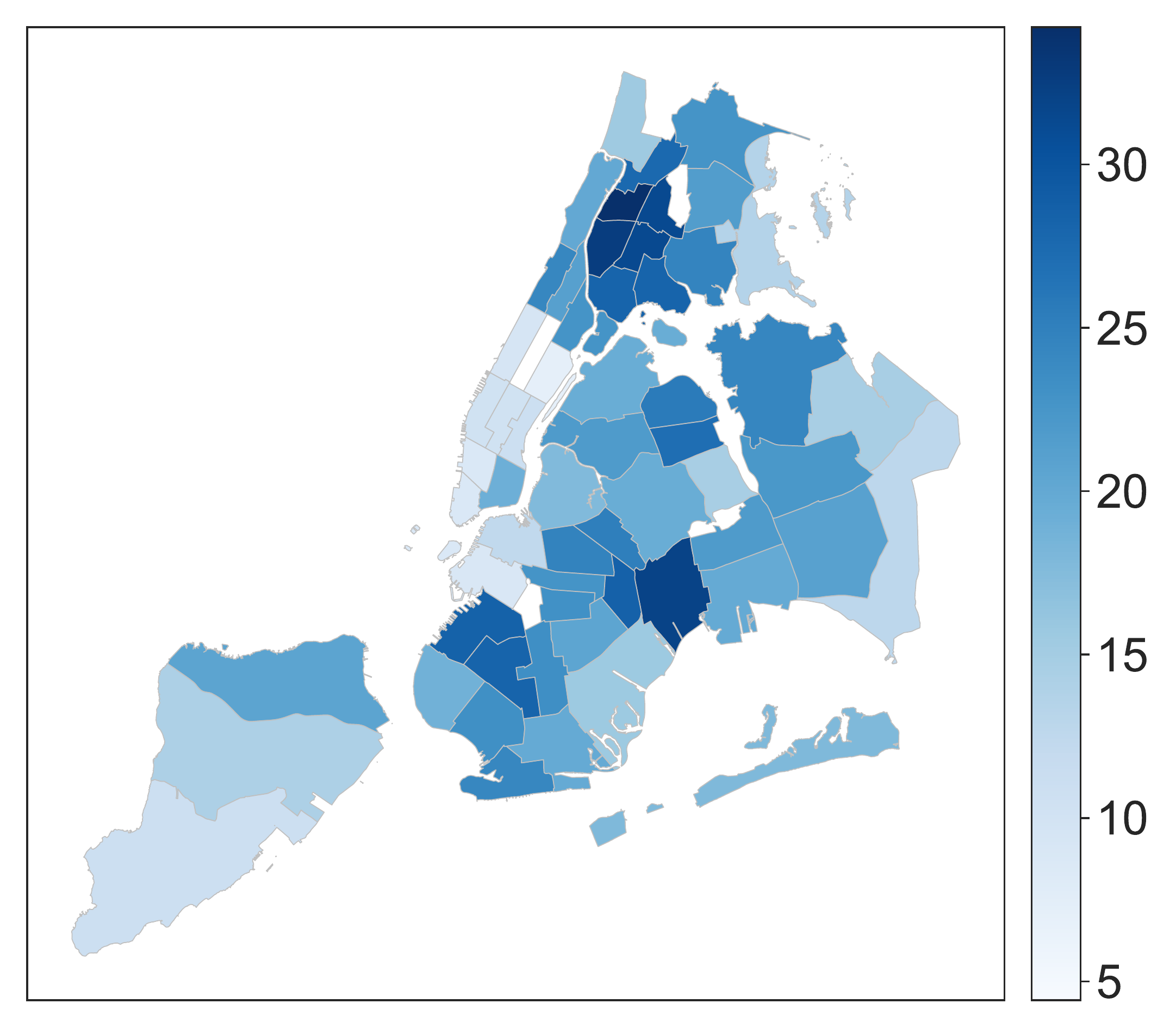}\label{fig:vis_areal_test}}
  \hspace{.2cm}
  \subfigure[A-GP]
  {\includegraphics[width=50mm, bb=0 0 622.9125 548.2649703756]{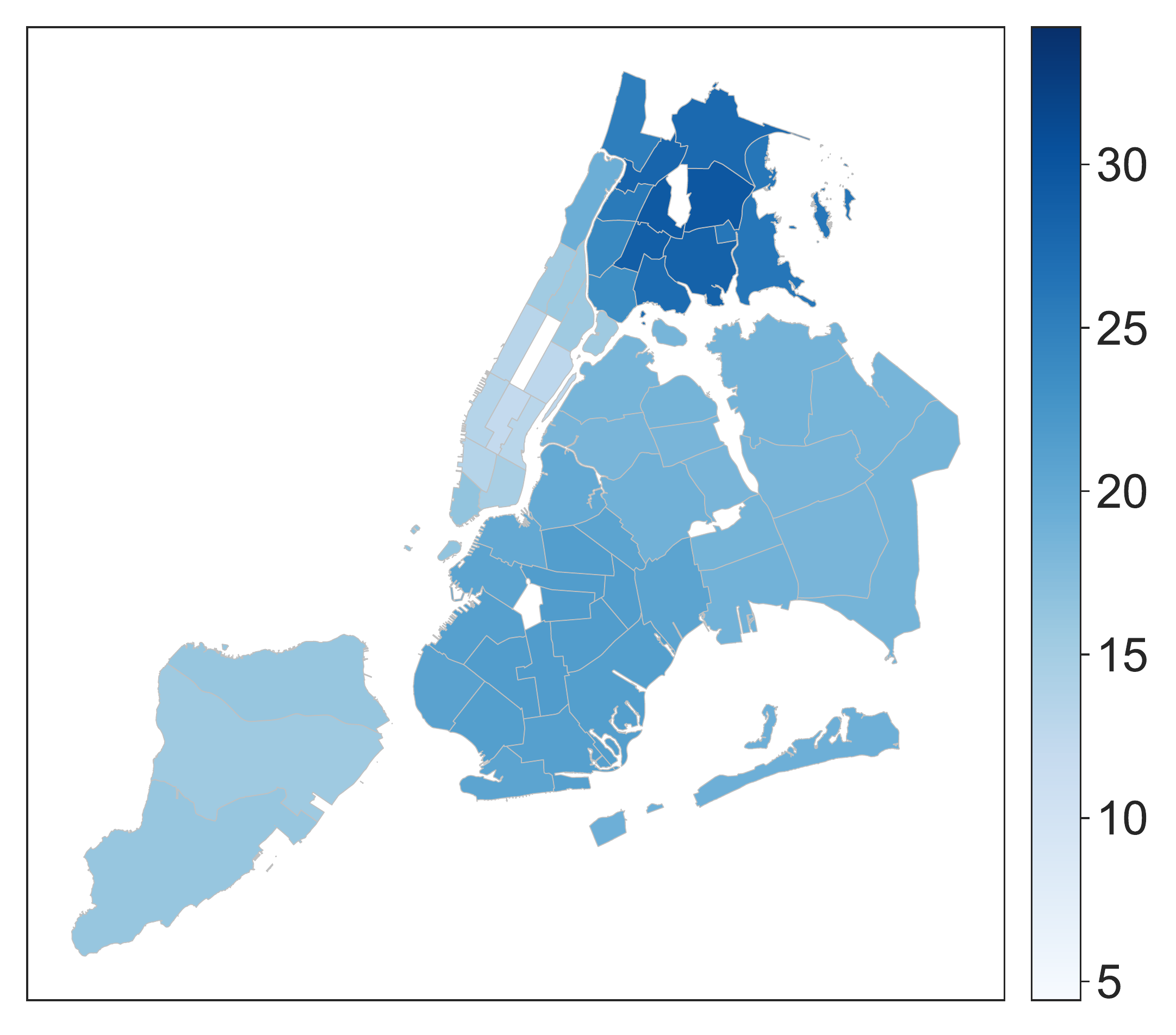}\label{fig:}}
  \hspace{.2cm}
  \subfigure[SLFM]
  {\includegraphics[width=50mm, bb=0 0 622.9125 548.2649703756]{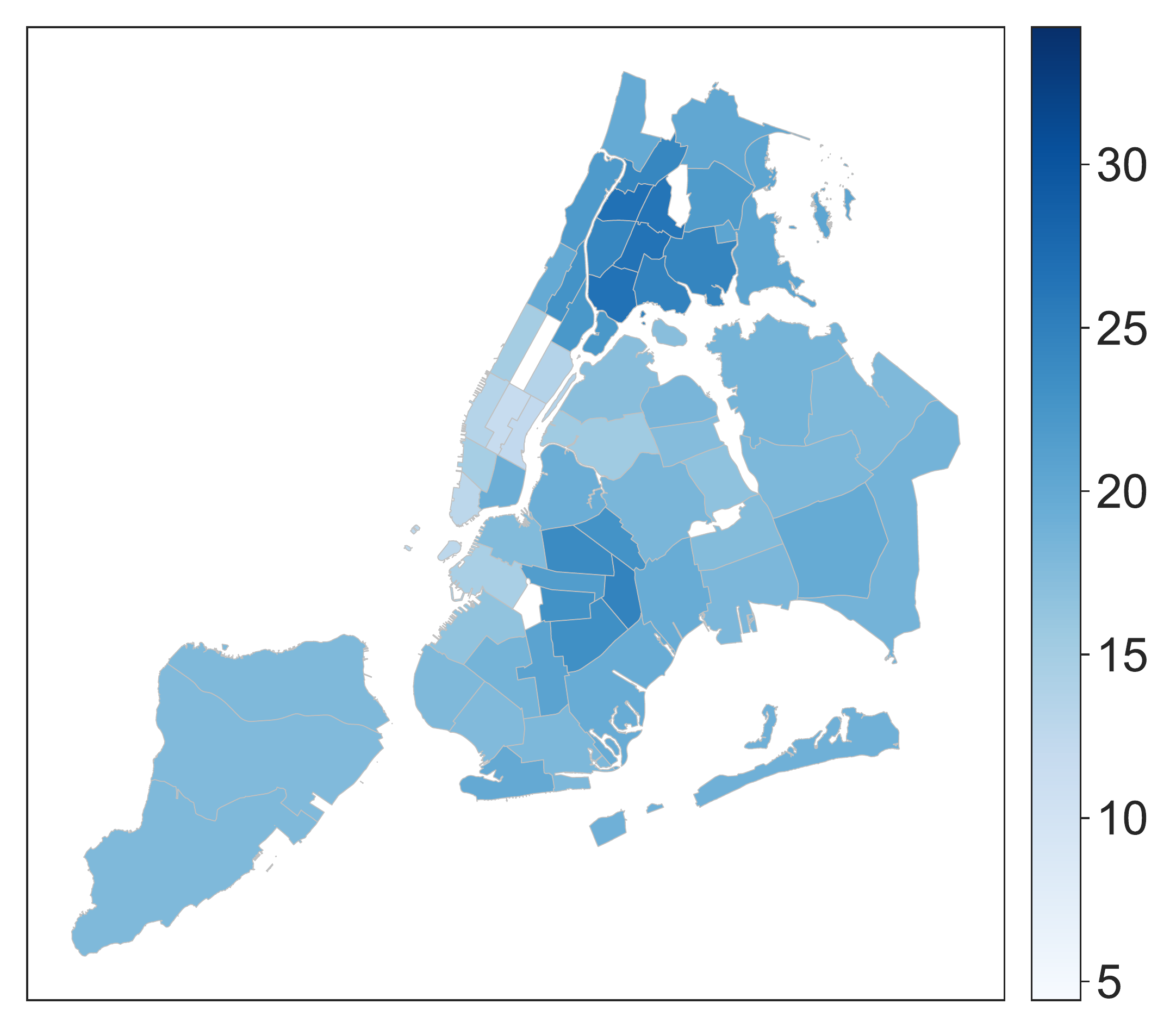}\label{fig:}}
  \hspace{.2cm}
  \subfigure[A-MoGP]
  {\includegraphics[width=50mm, bb=0 0 622.9125 548.2649703756]{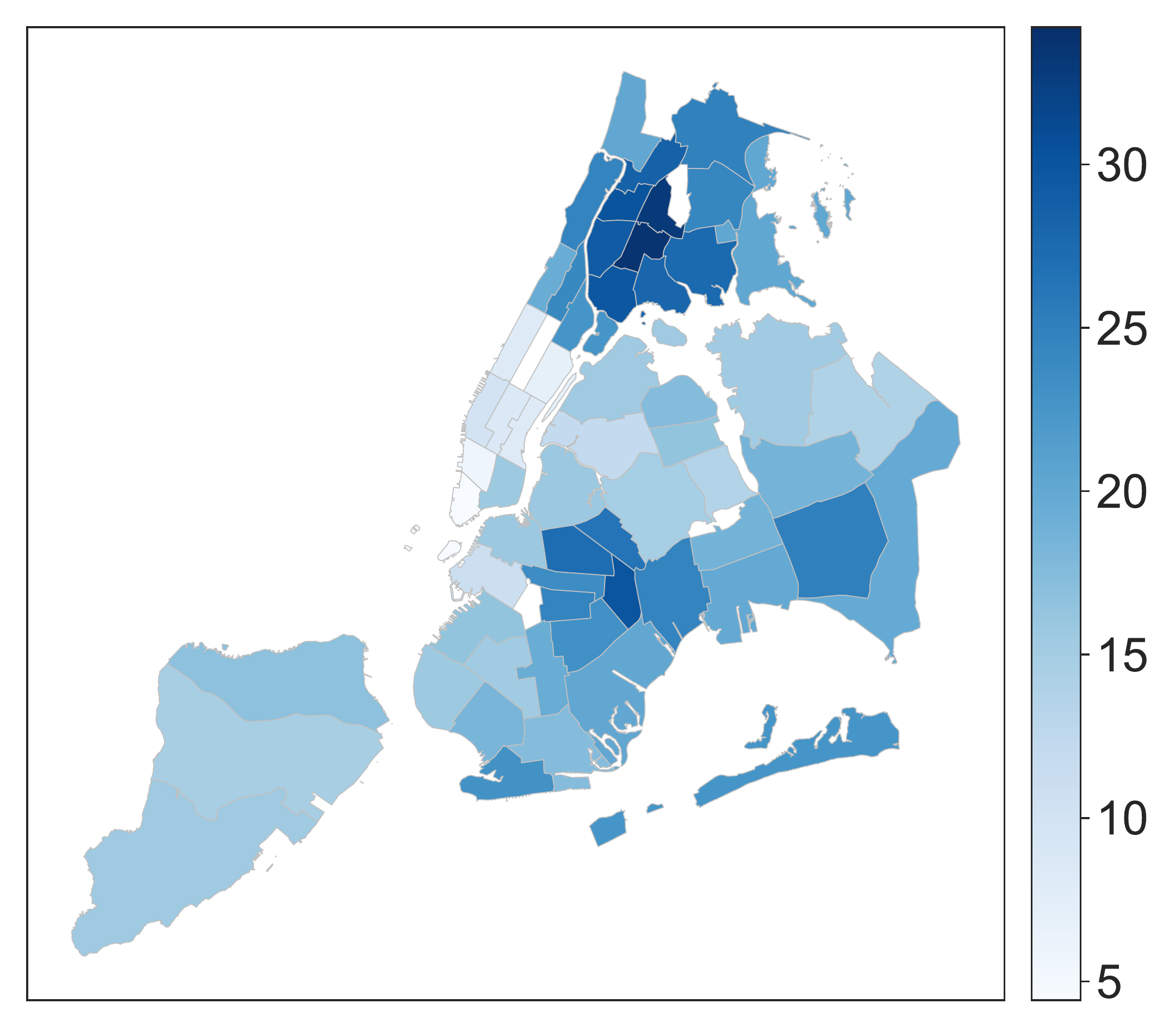}\label{fig:}}
  \hspace{.2cm}
  \subfigure[A-MoGP (trans)]
  {\includegraphics[width=50mm, bb=0 0 622.9125 548.2649703756]{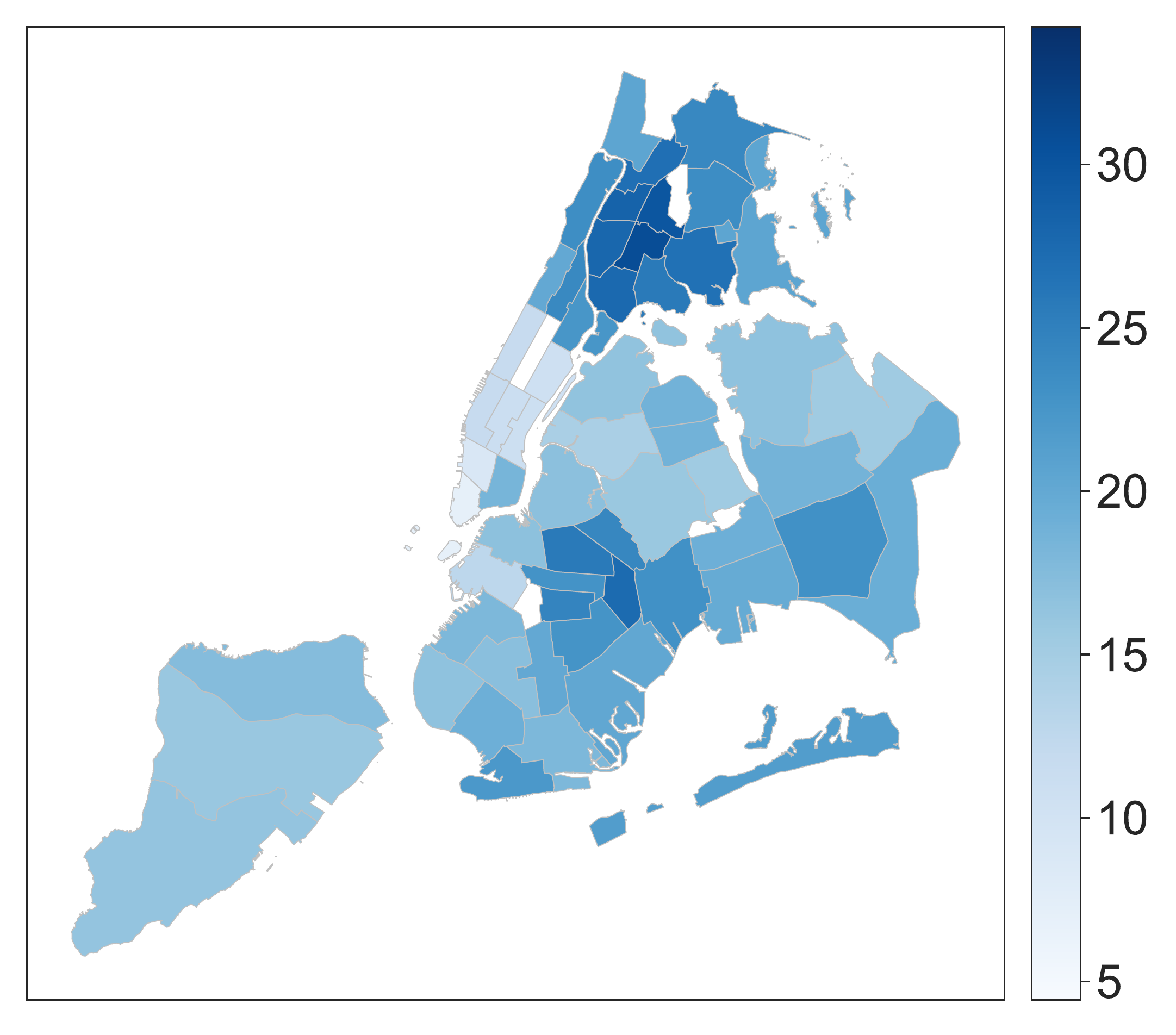}\label{fig:}}
 \end{center}
 \caption{Refinement results of the Poverty rate dataset in New York City.}
 \label{fig:vis_areal}
\end{figure*}

Figures~\ref{fig:vis_A-GP}, \ref{fig:vis_SLFM}, \ref{fig:vis_A-MoGP},
and \ref{fig:vis_A-MoGP(trans)}
illustrate the prediction results of the attributes in Aotizhongxin,
by A-GP, SLFM, A-MoGP, and A-MoGP (trans), respectively.
Here, the target dataset is set to NO2 in Aotizhongxin.
In these figures,
the first row is the result for the test dataset
(i.e., fine-grained target dataset).
The remaining rows in these figures are the results
for the training datasets,
including the coarse-grained target dataset.
In the results of A-GP, A-MoGP, and A-MoGP (trans),
the predictive variance (depicted by the red shaded area) was obtained
via the double integral of the posterior covariance function
$\hat{{\bf K}}_{v'}^*(\bm{x},\bm{x}')$~\eqref{eq:VB_post_cov}
over the corresponding bin.
Each column in these figures shows the result
when using the coarse-grained target dataset with the corresponding bin size
(described by the bold font in the figures).
As one can see from Figure~\ref{fig:vis_A-GP},
A-GP incorporating the aggregation process
can effectively interpolate the target aggregated attribute
without over-fitting
while encouraging consistency
between the coarse- and the fine-grained target dataset.
However, the performance of A-GP is limited
because it does not use multiple datasets
to refine the coarse-grained target dataset.
Although SLFM can utilize multiple datasets from a single domain,
it does not have the observation model
for handling aggregate datasets;
this might prevent the model from learning properly. 
In the third column of Figure~\ref{fig:vis_SLFM},
one observes the over-fitting of SLFM to the CO dataset
because SLFM regarded the areal data
as the point-referenced data.
It is one of the reasons why 
SLFM does not yield better performance
in refining the coarse-grained dataset.
As one can see from Figure~\ref{fig:vis_A-MoGP},
A-MoGP captured the fine-scale variation of the target attribute
by utilizing the relevant attribute (i.e., CO),
only in the case where
the target dataset has the relatively fine granularities (i.e., one month).
In Figure~\ref{fig:vis_A-MoGP(trans)},
one can see that A-MoGP (trans) yielded an excellent prediction
even if there is only the target dataset
with coarser granularities.
By comparing the estimated function
(depicted by the red lines in Figures~\ref{fig:vis_A-GP}, \ref{fig:vis_SLFM}, \ref{fig:vis_A-MoGP},
and \ref{fig:vis_A-MoGP(trans)})
when setting the target coarse-grained bin size
to four months among all the models,
we can see that the improvement with A-MoGP (trans) is obvious.
This is because A-MoGP (trans) can utilize
the aggregate datasets in multiple domains
and appropriately learn
the relationships between attributes
and the covariances of data points.
To confirm this observation, in Figure~\ref{fig:relationship},
we show the visualization of the relationships between attributes,
estimated by A-MoGP and A-MoGP (trans).
Letting $\bar{{\bf W}}'_v$ be an $|\mathcal{S}_v| \times |\mathcal{L}|$ matrix
whose $(s,l)$-entry is the variational posterior mean
$\bar{w}_{vsl}'$ in~\eqref{eq:variational_dist},
the relationships between attributes were calculated as follows:
$\bar{{\bf W}}'_v \bar{{\bf W}}'^{\top}_v \in \mathbb{R}^{|\mathcal{S}_v|\times|\mathcal{S}_v|}$,
which is known as a {\em coregionalization matrix}~\cite{alvarez:kernels}.
Figure~\ref{fig:relationship} illustrates the absolute values
of the elements of the coregionalization matrix.
This result shows that A-MoGP captured the useful relationship between attributes,
that is, the correlation between NO2 and CO,
only in the case where the bin size for the coarse-grained target attribute was one month.
Meanwhile, A-MoGP (trans) emphasized this relationship
even if the target attribute was associated with coarser time bins.

\begin{figure}[tb]
 \begin{center}
  \subfigure[Estimated relationships (Chicago).]
  {\includegraphics[width=55mm, bb=0 0 340.66971 186.697725]{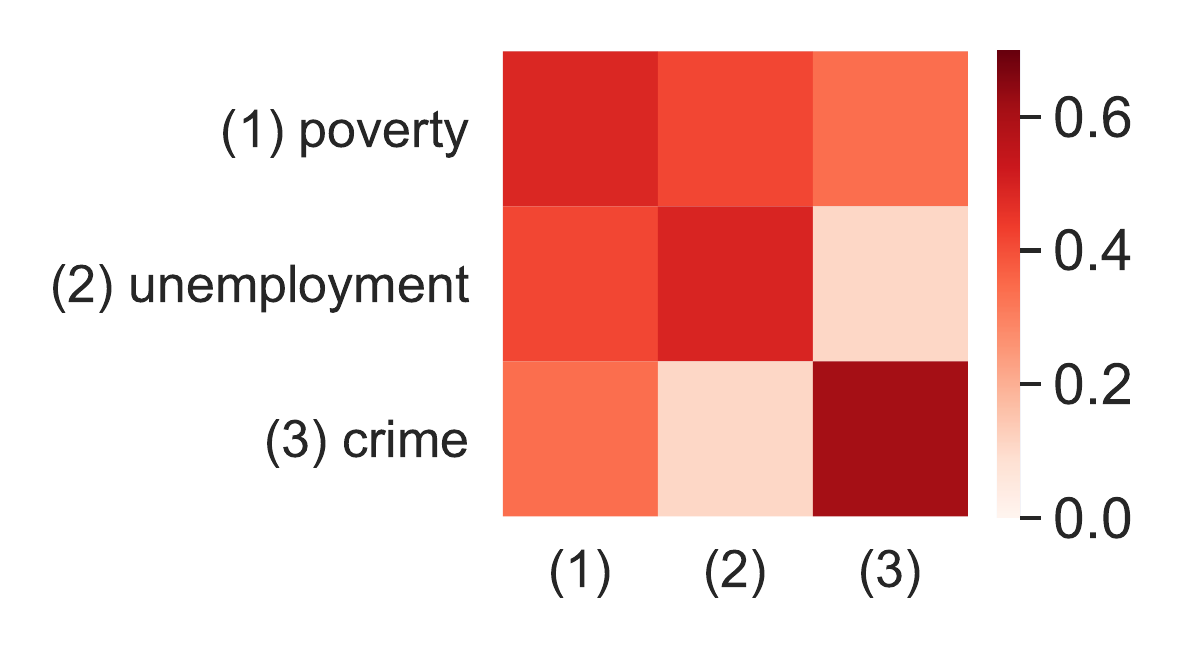}\label{fig:}}
  \subfigure[Estimated relationships (New York City).]
  {\includegraphics[width=85.8mm, bb=0 0 536.87099 369.196775]{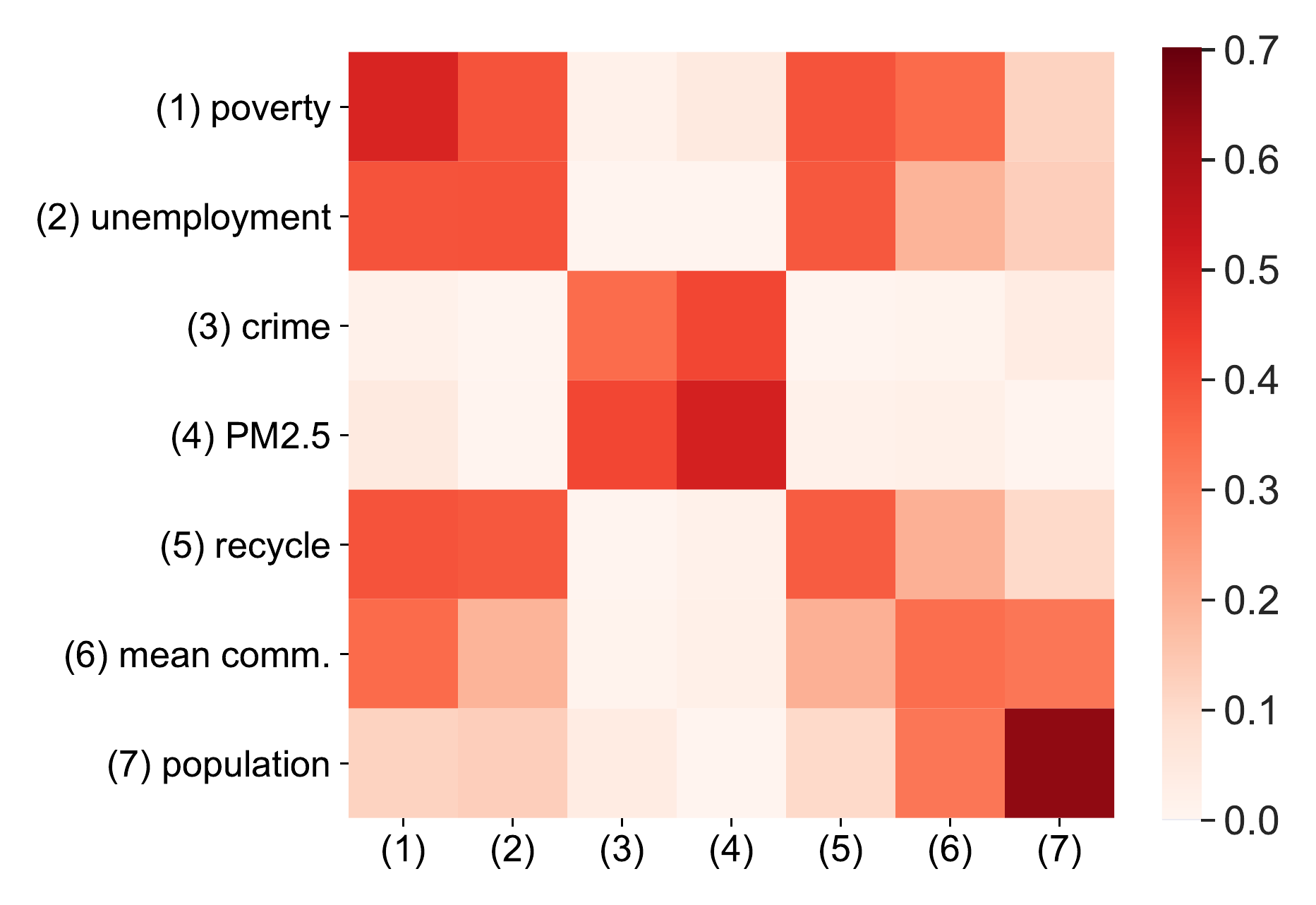}\label{fig:}}
 \end{center}
 \caption{Relationships between attributes learned by A-MoGP (trans).
 The target dataset was set to the Poverty rate dataset in New York City.}
 \label{fig:2d_relationship}
\end{figure}
\begin{figure}[tb]
 \centering
 \includegraphics[width=85mm, bb=0 0 1850 1302]{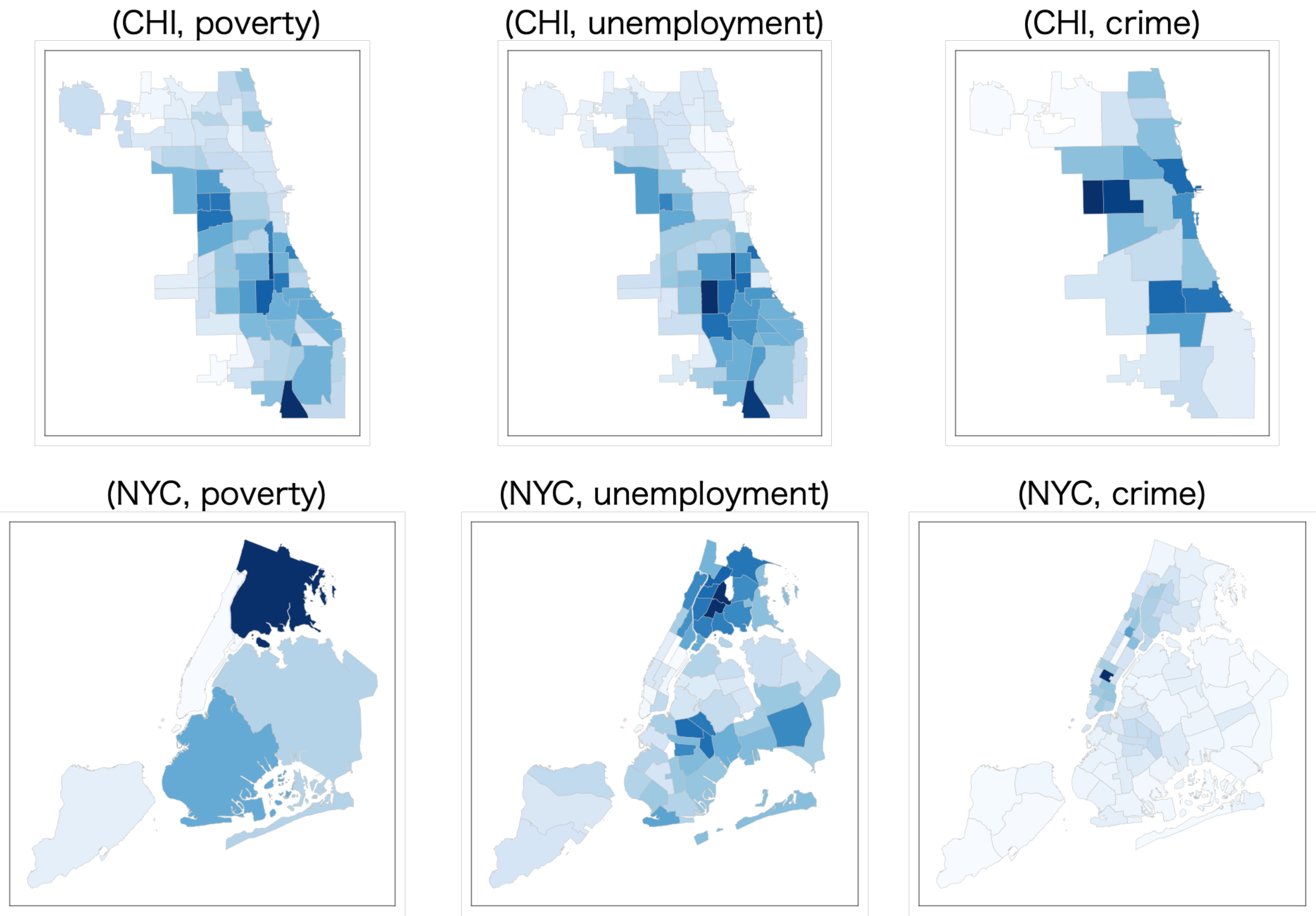}
 \caption{Visualizations of the training datasets contained in both Chicago and New York City.}
 \label{fig:vis_areal_datasets}
\end{figure}
\begin{figure}[tb]
 \centering
 \includegraphics[width=90mm, bb=0 0 1440 1520]{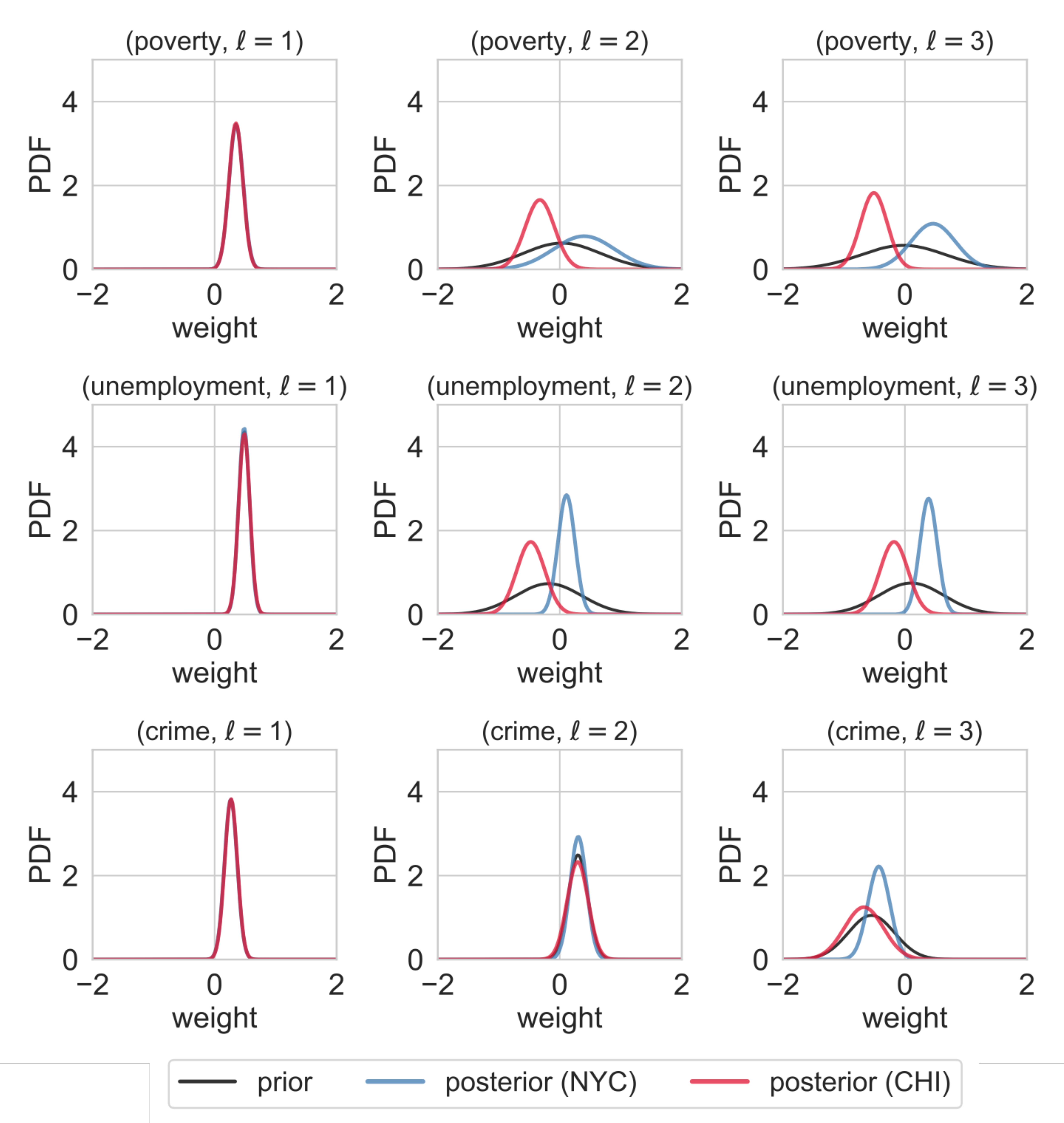}
 \caption{Visualizations of the prior and the variational posterior for weight parameters,
 estimated by A-MoGP (trans).}
 \label{fig:dist}
\end{figure}

\subsection{Results for areal data}
\label{sec:spatial}
This section presents the experimental results
for areal datasets in cities.
Table~\ref{tb:result_2d} shows MAPE and standard errors
for A-GP, SLFM, A-MoGP, and A-MoGP (trans).
Here, the experiments for the Crime rate dataset in Chicago
have not been conducted
because the coarser version for training
is not available online.
For all datasets,
A-MoGP and A-MoGP (trans) achieved better performance
in refining the coarse-grained areal dataset
than the baseline models.
Also, in most cases,
A-MoGP (trans) was able to use the aggregate datasets
in two cities to improve refinement performance.
Figure~\ref{fig:vis_areal} shows the Poverty rate dataset in New York City
and the refinement results attained by A-GP, SLFM, A-MoGP, and A-MoGP (trans).
Compared with the true values in Figure~\ref{fig:vis_areal_test},
A-MoGP (trans) yielded more accurate estimates of the fine-grained data than the other models.

We analyzed the relationships between the attributes,
estimated by A-MoGP (trans).
Figure~\ref{fig:2d_relationship} shows the visualization of the relationships
when setting the Poverty rate dataset in New York City
as the target dataset,
which was illustrated by the same procedure
as that described in Section~\ref{sec:1d_result}.
In Figure~\ref{fig:2d_relationship},
one can see that the poverty rate and the unemployment rate
had strong relationships in both cities;
meanwhile, the strength of the relationship
between the poverty rate and the crime rate
was stronger in Chicago than that in New York City.
Comparing these results
with the visualizations of training datasets
in Figure~\ref{fig:vis_areal_datasets},
we can confirm that A-MoGP (trans) captures
the correlations between three attributes
(i.e., poverty rate, unemployment rate, and crime rate) appropriately.
One advantage of A-MoGP (trans) is that
it has the mechanism for knowledge transfer across domains,
allowing for utilizing datasets from multiple domains
to learn weight parameters
defining relationships between attributes.
Sharing the prior distribution~\eqref{eq:prior_w}
for the weight parameters among all domains
makes the knowledge transfer possible;
Figure~\ref{fig:dist} visualizes
the prior~\eqref{eq:prior_w} and
the variational posterior~\eqref{eq:variational_dist},
estimated by A-MoGP (trans).
Each tile shows the distribution for the weight $w_{vsl}$
of the $l$-th latent GP in each attribute.
One observes that the prior variances
were small in the case of $l=1$
and in the case of the Crime rate dataset of $l=2$
(see black lines in Figure~\ref{fig:dist});
the variational posteriors for New York City
(blue lines in Figure~\ref{fig:dist})
and Chicago (red lines in Figure~\ref{fig:dist})
were close to the prior distributions.
In these cases,
knowledge transfer is more encouraged.
Accordingly,
Table~\ref{tb:result_2d} and Figure~\ref{fig:dist}
show that A-MoGP (trans) achieves performance improvement
by using the aggregate datasets from both cities
via the knowledge transfer.

\section{Conclusion}
\label{sec:conclusion}
In this article,
we have proposed the Aggregated Multi-output Gaussian Process (A-MoGP)
that can estimate functions for attributes
by utilizing aggregate datasets of respective granularities
from multiple domains.
Experiments on real-world datasets have confirmed that
the A-MoGP can perform better than baselines
in refining coarse-grained aggregate data;
moreover, the A-MoGP has improved performance
by using aggregate datasets from multiple domains
via the knowledge transfer across domains.

There are several research directions
that can be explored in the future.
First, we can introduce alternative likelihoods
such as the Poisson distribution for count data.
Second, we can speed up the inference
using the inducing point technique, similarly to~\cite{titsias:variational}.
Third, incorporating rich additional data is promising
to improve refinement performance.
For example, satellite images are known to be helpful
in the refinement of spatially aggregated data~\cite{Law:variational}.
A powerful option to leverage such data
is to use deep neural networks as a model component,
which would help automatically extract meaningful representations from data.


%

\ifCLASSOPTIONcaptionsoff
  \newpage
\fi

\end{document}